\documentclass[11pt]{article}

\usepackage[preprint]{acl}

\usepackage{times}
\usepackage{latexsym}

\usepackage[T1]{fontenc}

\usepackage[utf8]{inputenc}

\usepackage{microtype}

\usepackage{inconsolata}

\usepackage{graphicx}

\title{Linear-Time and Constant-Memory Text Embeddings\\ Based on Recurrent Language Models}

\author{Tobias Grantner \\
  Dynatrace Research \\
  \footnotesize{\texttt{tobias.grantner@dynatrace.com}}
  \And
  Emanuel Sallinger \\
  TU Wien \& University of Oxford \\
  \footnotesize{\texttt{emanuel.sallinger@tuwien.ac.at}}
  \And
  Martin Flechl \\
  Dynatrace Research \\
  \footnotesize{\texttt{martin.flechl@dynatrace.com}}
}

\usepackage{booktabs}

\usepackage{multirow}

\usepackage{rotating}

\usepackage{makecell}

\usepackage{import}

\usepackage{amsmath}

\usepackage{bm}

\usepackage{amsfonts}

\usepackage{todonotes}

\usepackage[inline]{enumitem}

\usepackage{adjustbox}

\usepackage{etoolbox}

\usepackage{keyval}

\usepackage{tabularx}

\usepackage{caption}
\usepackage{subcaption}

\usepackage{float}

\renewcommand*{\vec}[1]{\bm{#1}}
\newcommand*{\mat}[1]{\bm{#1}}
\newcommand*{\languagemodel}{\mathcal{M}}
\newcommand*{\embeddingmodel}{\mathcal{E}}
\newcommand*{\token}{t}
\newcommand*{\sequence}{\vec{s}}
\newcommand*{\embedding}{\vec{e}}
\newcommand*{\similarity}[2]{\cos(#1, #2)}
\newcommand*{\temperature}{\tau}
\newcommand*{\pexp}[1]{e^{#1}}

\newcommand*{\R}{\mathbb{R}}
\newcommand*{\A}{\mat{A}}
\newcommand*{\B}{\mat{B}}
\newcommand*{\C}{\mat{C}}
\newcommand*{\M}{\mat{M}}
\newcommand*{\matL}{\mat{L}}
\newcommand*{\smalldiag}[1]{\operatorname{diag}(#1)}
\newcommand*{\diag}[1]{\operatorname{diag}\bigl(#1\bigr)}
\newcommand*{\I}{\mat{I}}
\newcommand*{\x}{\vec{x}}
\newcommand*{\y}{\vec{y}}
\newcommand*{\h}{\vec{h}}
\newcommand*{\hb}{\vec{b}}
\newcommand*{\zerovec}{\vec{0}}
\newcommand*{\chunked}[1]{\widehat{#1}}

\newcommand*{\modelefive}{\texorpdfstring{E5\textsubscript{mistral-7b}}{E5-mistral-7b}}
\newcommand*{\modelqwen}{Qwen2~1.5B}
\newcommand*{\modelmistral}{Mistral~7B}
\newcommand*{\modelmamba}{Mamba2~1.3B}
\newcommand*{\modelcodestral}{Codestral~Mamba2~7B}
\newcommand*{\modelcodestralshort}{Cod.~Mamba2~7B}
\newcommand*{\modelrwkv}{RWKV7~7.2B}
\newcommand*{\modelxlstm}{xLSTM~7B}

\newcommand*{\version}[1]{\texttt{#1}}
\newcommand*{\modelefivewithversion}{\texorpdfstring{E5\textsubscript{mistral-7b (\version{v0.1})}}{E5-mistral-7b (\version{v0.1})}}
\newcommand*{\modelmistralversion}[1]{\modelmistral~\version{#1}}

\newcommand*{\mteb}{MTEB}
\newcommand*{\englishmtebold}{\texttt{MTEB(eng,~v1)}}
\newcommand*{\englishmteb}{\texttt{MTEB(eng,~v2)}}
\newcommand*{\multilingualmteb}{\texttt{MTEB(Multilingual,~v2)}}
\newcommand*{\longembed}{\texttt{LongEmbed}}

\newcommand*{\hflink}[1]{  \hspace{1.5px}  \adjustbox{valign=m}{\includegraphics[height=1em]{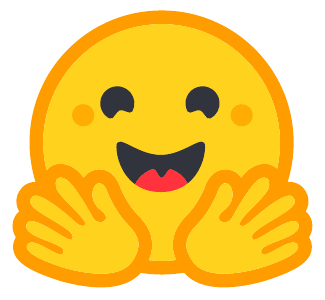}}  \hspace{3px}  \adjustbox{valign=m}{\href{https://huggingface.co/#1}{\texttt{#1}}}}

\newcommand{\expnumber}[2]{{#1} \cdot 10^{#2}}

\newcommand*{\bigO}[1]{\mathcal{O}(#1)}

\begin{document}
\maketitle
\begin{abstract}
Transformer-based embedding models suffer from quadratic computational and linear memory complexity, limiting their utility for long sequences. We propose recurrent architectures as an efficient alternative, introducing a vertically chunked inference strategy that enables fast embedding generation with memory usage that becomes constant in the input length once it exceeds the vertical chunk size. By fine-tuning Mamba2 models, we demonstrate their viability as general-purpose text embedders, achieving competitive performance across a range of benchmarks while maintaining a substantially smaller memory footprint compared to transformer-based counterparts. We empirically validate the applicability of our inference strategy to Mamba2, RWKV, and xLSTM models, confirming consistent runtime-memory trade-offs across architectures and establishing recurrent models as a compelling alternative to transformers for efficient embedding generation.
\end{abstract}

\section{Introduction}
Text embeddings are central to modern natural language processing, enabling tasks such as information retrieval, clustering, and semantic similarity estimation. While transformer-based embedding models represent the state of the art, their quadratic computational and linear memory complexity in the input length restricts their use for long documents and in resource-constrained environments.

Recent advances in recurrent architectures such as Mamba, RWKV, and xLSTM \citep{gu_mamba_2024,dao_transformers_2024,peng-etal-2023-rwkv,peng_rwkv_2025,beck_xlstm_2024} achieve competitive language modeling performance while offering linear computational scaling and constant memory usage. Despite extensive work on the efficiency of these architectures for text generation, their use for general-purpose embedding generation remains largely unexplored, with existing work limited to small task-specific models \citep{zhang_mamba_2024,liang_cognitive-inspired_2025,pan_exploring_2025,cao_single-pass_2025}.

We investigate recurrent architectures as efficient alternatives to transformer-based text embedding models by applying the \modelefive\ training procedure \citep{wang-etal-2024-improving-text} to pretrained Mamba2-based language models \citep{dao_transformers_2024}. Our results on the English Massive Text Embedding Benchmark (\mteb), Multilingual \mteb, and the \longembed\ benchmark \citep{muennighoff-etal-2023-mteb,enevoldsen_mmteb_2025,zhu-etal-2024-longembed} demonstrate competitive performance against strong transformer baselines, validating the viability of recurrent models as general-purpose text embedders.

Beyond architectural adaptation, we leverage the duality of Mamba2 to derive matrix formulations for chunked inference, building on \citet{dao_transformers_2024}, that enable parallelization across chunks while maintaining linear computational complexity. We then propose a vertically chunked inference strategy that processes inputs in fixed-size blocks across model layers, balancing parallelization with recurrent processing to keep memory usage constant in the input length. This strategy applies broadly to recurrent architectures with a linear-time recurrence and a complementary parallelizable formulation for intra-chunk processing. We validate it empirically on Mamba2, RWKV, and xLSTM models. Our results show that parallelization benefits saturate at small vertical chunk sizes across all three architectures, yielding favorable memory efficiency over transformers without considerable runtime overhead, particularly for long documents.

We release our fine-tuned Mamba2-based model checkpoints together with our inference implementation, to support reproducibility and encourage further research.\footnote{\hflink{collections/dynatrace-oss/embed-mamba2}}

\section{Background}

\subsection{Embedding Model Training}
The transformer architecture has become the de facto standard for text embedding models. Early approaches adapted masked language models like BERT \citep{devlin-etal-2019-bert} through multi-stage weak supervision. Following \modelefive\ \citep{wang-etal-2024-improving-text}, decoder-only models have shifted the paradigm by fine-tuning autoregressive language models on diverse synthetic and annotated data. Despite further advancements driven by improved synthetic data generation, stronger foundation models, and extended training procedures \citep{zhang_qwen3_2025,choi_linq-embed-mistral_2024,vera_embeddinggemma_2025}, \modelefive\ remains competitive among general-purpose open-weight embedding models.

\subsection{Efficient Embedding Inference}
Efficiency gains in transformer inference typically rely on attention optimizations or quantization. FlashAttention \citep{dao_flashattention_2022,dao_flashattention-2_2024,shah_flashattention-3_2024} uses tiling to avoid the materialization of the quadratic attention matrix, reducing the memory footprint of the attention computation from quadratic to linear in the sequence length while improving runtime through increased hardware utilization. Sliding-window attention \citep{child_generating_2019,jiang_mistral_2023} restricts the context to a fixed local window, achieving linear computational complexity at the cost of expressiveness. Quantization techniques reduce the memory footprint of model weights and activations by representing them in lower-precision formats. While static embedding models \citep{tulkens_model2vec_2024} offer the highest efficiency and constant memory usage, they generally lag behind transformer-based models in embedding quality.

\subsection{Recurrent Language Models}
Recurrent neural networks (RNNs) scale linearly with the input length by maintaining a constant-sized latent state. Traditional RNNs lack the parallelizability required for large-scale training, but modern variants like Mamba2 \citep{dao_transformers_2024}, RWKV \citep{peng-etal-2023-rwkv,peng_rwkv_2025}, and xLSTM \citep{beck_xlstm_2024} introduce structured recurrence mechanisms that admit parallel computation of state updates. Although their text generation capabilities have been extensively studied, their potential as general-purpose text embedders has received comparatively little attention.

\section{Recurrent Embedding Models}
\label{sec:recurrent-embedding-models}

\subsection{Training}
\label{sec:recurrent-embedding-models:training}
To evaluate the effectiveness of embedding models based on recurrent language model architectures, we adopt the contrastive training recipe established by \modelefive\ \citep{wang-etal-2024-improving-text}. Given a pretrained model $\languagemodel$, we adapt it for embedding generation by removing the language modeling head to create $\languagemodel'$. For each input sequence $\sequence$ consisting of tokens $(\token_1, \ldots, \token_n) \in \mathcal{T}^n$, we append the model's end-of-sequence (EOS) token $\token^{\mathtt{EOS}}$. The output of the final model layer at this terminal position $\embedding^{\mathtt{EOS}} \in \R^d$ serves as a summary of the preceding context, which we use as the representation $\embeddingmodel(\sequence)$ of the input text:
\begin{equation}
\begin{split}
  \sequence &= (\token_1, \token_2, \ldots, \token_n) \\
  \sequence \oplus \token^{\mathtt{EOS}} &= (\token_1, \token_2, \ldots, \token_n, \token^{\mathtt{EOS}}) \\
  \languagemodel'(\sequence \oplus \token^{\mathtt{EOS}}) &= (\embedding_1, \embedding_2, \ldots, \embedding_n, \embedding^{\mathtt{EOS}}) \\
  \embeddingmodel(\sequence) &= \embedding^{\mathtt{EOS}}
\end{split}
\end{equation}

\noindent
The resulting model $\embeddingmodel : \mathcal{T}^* \to \R^d$ maps a variable-length text sequence to a $d$-dimensional real-valued output of fixed size. We fine-tune it for embedding generation using the InfoNCE loss, which encourages the model to distinguish a relevant positive pair from a set of negative examples \citep{oord_representation_2019}:
\begin{equation}
  \mathcal{L} = - \log \frac{\pexp{\similarity{\embedding_q}{\embedding_p} / \temperature}}{\pexp{\similarity{\embedding_q}{\embedding_p} / \temperature} + \sum_{i=1}^{N} \pexp{\similarity{\embedding_q}{\embedding_{n_i}} / \temperature}}
\end{equation}

\noindent
Here, $\embedding_q$ and $\embedding_p$ represent the embeddings of the query and the positive sample, respectively, $\similarity{\cdot}{\cdot}$ is the cosine similarity function, and $\temperature$ is a temperature hyperparameter. The training data consists of query-positive pairs, along with zero, one, or multiple hard negative examples. The $N$ in-batch negatives $\embedding_{n_i}$ comprise both the hard negatives provided with the query, and all positives and hard negatives associated with the other queries in the batch.

Following \citet{wang-etal-2024-improving-text}, we incorporate instruction tuning to enable the model to adapt to varied downstream tasks without task-specific training. We use the following template for all queries:
\begin{quote}
  \texttt{Instruction:\,\textvisiblespace\,\{prompt\}\,\textbackslash n\,Query:\,\textvisiblespace}
\end{quote}

\noindent
where \texttt{\{prompt\}} is replaced with the task-specific instruction, and \texttt{\textbackslash n} is a newline character. By prepending this instruction only to the query, not to the documents, we ensure the system remains efficient for large-scale retrieval, since the document corpus only needs to be embedded once, regardless of the specific task instruction.

\subsection{Recurrent Embedding Inference}
\label{sec:recurrent-embedding-models:inference}
While the inference efficiency of recurrent language models during autoregressive generation is well-documented, their behavior during the initial encoding of input tokens, which constitutes the inference process for embedding generation, remains underexplored.

We investigate recurrent embedding inference using Mamba2 as a representative architecture because its dual nature allows it to be formulated as both a linear recurrence and a structured matrix transformation. We propose strategies for optimizing parallelization levels and memory scheduling. Conceptually, our scheduling strategy requires two properties: \begin{enumerate*}[label=\roman*)] \item a linear-time recurrent structure over the input sequence and \item a parallelizable dual formulation that supports intra-chunk processing in addition to the recurrent mode.\end{enumerate*} Because these properties are shared by a broader family of recurrent and SSM-style architectures, including RWKV and xLSTM (see Appendix~\ref{app:recurrent-architectures}), our strategy is widely applicable. We empirically validate this in Section~\ref{sec:experiments:inference-evaluation} on all three architectures, while focusing on Mamba2 for the formal derivation. For clarity, our discussion omits discretization and focuses on single-dimensional input and output. The concepts generalize independently to multiple dimensions, and we refer to \citet{dao_transformers_2024} for proofs and the extension to discretized, multi-dimensional input.

\subsubsection{The Duality of Mamba2}
Mamba2 \citep{dao_transformers_2024} is a structured state space model (SSM) that maps an input sequence $\x \in \R^T$ to an output sequence $\y \in \R^T$ of length $T$. The output $\y_t$ at each time step $t \in \{1, \ldots, T\}$ depends on the current input $\x_t$ and a latent state vector $\h_{t-1} \in \R^N$ with a state expansion factor $N$. The state is updated iteratively, and the output is generated as follows:
\begin{subequations}
  \label{eq:mamba2}
  \begin{align}
  \label{eq:mamba2:state}
    \h_{t} &= \A_t \h_{t-1} + \B_t \x_t \\
  \label{eq:mamba2:output}
    \y_t &= \C_t^{\top} \h_t
  \end{align}
\end{subequations}

\noindent
The time-variant matrices $\A \in \R^{T \times N \times N}$, $\B \in \R^{T \times N}$ and $\C \in \R^{T \times N}$ are functions of the input, typically computed via a combination of learned projections and non-linear transformations. This recurrent formulation enables inference with $\bigO{T}$ computational complexity and $\bigO{1}$ memory usage with respect to the sequence length.
However, its sequential computation of latent state updates prevents it from fully exploiting the parallelization capabilities of modern hardware accelerators.

To bridge this gap, structured SSMs constrain the transition matrix $\A_t$ to allow for efficient computation of state updates. For Mamba2, $\A_t = a_t \I$, where $a_t$ is a scalar and $\I$ corresponds to the identity matrix. This structure allows the recurrence in Equation~\ref{eq:mamba2:state} to be unrolled. Assuming an initial state $\h_0 = \zerovec$, the state at any time $t$ (and consequently the output $\y_t$) can be expressed as a direct function of all preceding inputs:
\begin{subequations}
  \begin{align}
    \A^\times_{i:j} &:= \begin{cases}
      \prod_{k=j+1}^{i} a_k \, \I, & i > j \\
      \I, & i = j \\
      \mat{0}, & i < j
    \end{cases} \\
    \begin{split}
      \y_t &= \sum_{s=1}^t \C_t^{\top} \A_t \cdots \A_{s+1} \B_s \x_s \\
      &= \sum_{s=1}^t \C_t^{\top} \A^\times_{t:s} \B_s \x_s
    \end{split}
  \end{align}
\end{subequations}
\noindent
Here, $\A^\times_{i:j}$ represents the cumulative transition from step $j$ to $i$, taking advantage of the structure $\A_t = a_t \I$. The entire sequence transformation can therefore be viewed as a single linear operator $\M$:\begin{subequations}
  \label{eq:ssm-matrix}
  \begin{align}
    \label{eq:ssm-matrix:M}
    \M_{i,j} &:= \C_i^{\top} \A^\times_{i:j} \B_j \\
    \label{eq:ssm-matrix:y}
    \y &= \M \x
  \end{align}
\end{subequations}

\noindent
As a consequence of the recurrence in Equation~\ref{eq:mamba2:state}, $\M$ is a lower-triangular matrix, which is encoded by the value of $\A^\times_{i:j}$ for $i < j$.
We can further vectorize the computation of $\M$ by defining a causal kernel matrix $\matL \in \R^{T \times T}$ with:
\begin{subequations}
  \label{eq:ssm-matrix-vectorized}
  \begin{align}    \label{eq:ssm-matrix-vectorized:L}
    \matL_{i,j} &:=
    \begin{cases}
    \prod_{k=j+1}^{i} a_k, & i > j \\
    1, & i = j \\
    0, & i < j
    \end{cases} \\
    \label{eq:ssm-matrix-vectorized:M}
    \M &= \matL \circ (\C \B^{\top})
  \end{align}
\end{subequations}

\noindent
This matrix formulation is key to the training efficiency of Mamba2, as it leverages parallelizable, hardware-accelerated matrix multiplications. However, instantiating the full matrix $\M$ incurs $\bigO{T^2}$ complexity in both computation and memory. The \emph{duality} of the model lies in its ability to switch between these two views, producing the same output: the fully parallelizable quadratic matrix formulation, and the memory-efficient linear recurrent view from Equation~\ref{eq:mamba2}.

\begingroup
\allowdisplaybreaks

\subsubsection{Chunked Inference}
\label{sec:recurrent-embedding-models:chunked-inference}

\begin{figure}[t]
  \includegraphics[width=\linewidth]{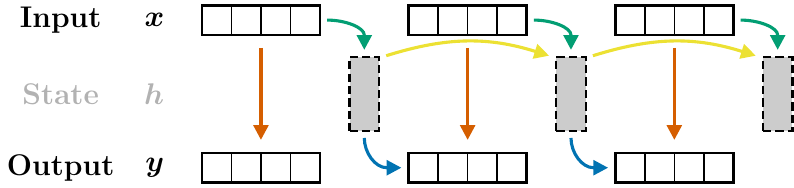}
  \caption{Chunked inference with parallel computation of chunks and recurrent state updates between chunks.}
  \label{fig:chunked-inference}
\end{figure}

The duality of Mamba2 enables the combination of the memory efficiency of recurrence with the parallelism of matrix multiplication. This is achieved by partitioning the input sequence into $K = T / Q$ chunks, each of size $Q$, assuming $Q$ divides $T$ for simplicity. Local chunk results are computed in parallel using the matrix formulation from Equation~\ref{eq:ssm-matrix}, while the global latent state is iteratively updated between chunks as illustrated in Figure~\ref{fig:chunked-inference}. Under this strategy, the dominant intra-chunk work is quadratic in the chunk size $Q$, yielding a total cost of $\bigO{K \cdot Q^2} = \bigO{T \cdot Q}$ over a sequence of length $T$.
By leveraging the semi-separable structure of $\M$, i.e., the property that its off-diagonal blocks admit low-rank factorizations through the shared transition matrices $\A$, we can additionally parallelize most of the computation across chunks through a block-decomposition matrix multiplication algorithm. A detailed derivation from first principles is provided in Appendix~\ref{app:chunked-inference-derivation}, while the key steps are outlined in this section.

We define chunked views of our previously introduced matrices and vectors
for chunk indices ${c, c' \in \{1, \ldots, K\}}$, within-chunk indices $i, j \in \{1, \ldots, Q\}$ and latent state indices $k, l \in \{1, \ldots, N\}$:
\begin{subequations}
  \begin{align}
    \chunked{\matL}^{(c,c')}_{i,j} &:= \matL_{(c-1) Q + i, (c'-1) Q + j} \\
    \chunked{\A}^{(c)}_{i,k,l} &:= \A_{(c-1) Q + i, k, l} \\
    \chunked{\A}^{(c)^\times}_{i:j} &:= \A^\times_{(c-1)Q+i:(c-1)Q+j} \\
    \chunked{\B}^{(c)}_{i,k} &:= \B_{(c-1) Q + i, k} \\
    \chunked{\C}^{(c)}_{i,k} &:= \C_{(c-1) Q + i, k} \\
    \chunked{\x}^{(c)}_{i} &:= \x_{(c-1) Q + i}
  \end{align}
\end{subequations}
\noindent
For any block $\chunked{\matL}^{(c,c')}$, we use $\chunked{\matL}^{(c,c')}_{i,\cdot} \in \R^Q$ to denote row $i$ and $\chunked{\matL}^{(c,c')}_{\cdot,j} \in \R^Q$ to denote column $j$.
To distinguish chunk-boundary states from token-level recurrent states, we define the boundary state after chunk $c$ by $\hb^{(c)} := \h_{cQ}$.
The computation is split into three stages:
\begin{enumerate*}[label=\roman*)]
  \item intra-chunk computation,
  \item inter-chunk state propagation, and
  \item the final output adjustment.
\end{enumerate*}

\paragraph{Intra-chunk Computation (Parallelizable)}\mbox{}

\begin{figure*}[t]
  \begin{subfigure}[b]{0.48\linewidth}
    \includegraphics[width=\linewidth]{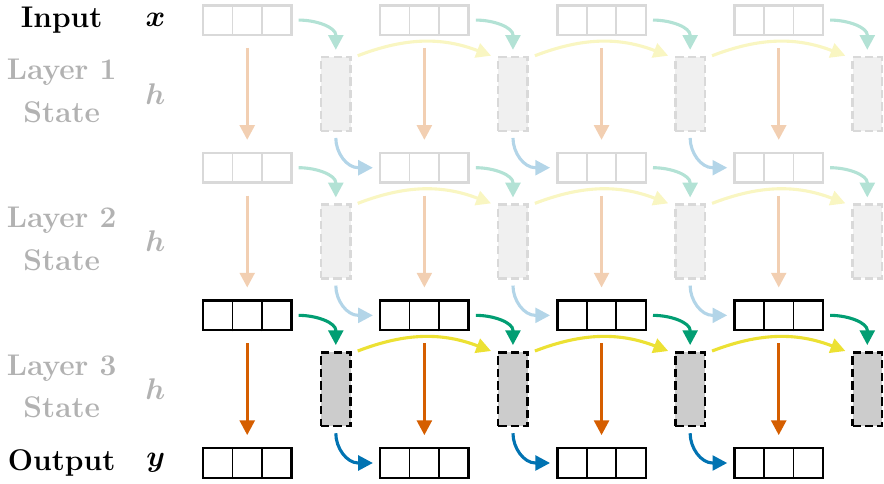}
    \caption{Fully horizontal inference: parallelization spans the full sequence length, but all intermediate activations for the currently processed layer must be retained in memory.}
    \label{fig:cross-layer-inference:horizontal}
  \end{subfigure}
  \hfill
  \begin{subfigure}[b]{0.48\linewidth}
    \includegraphics[width=\linewidth]{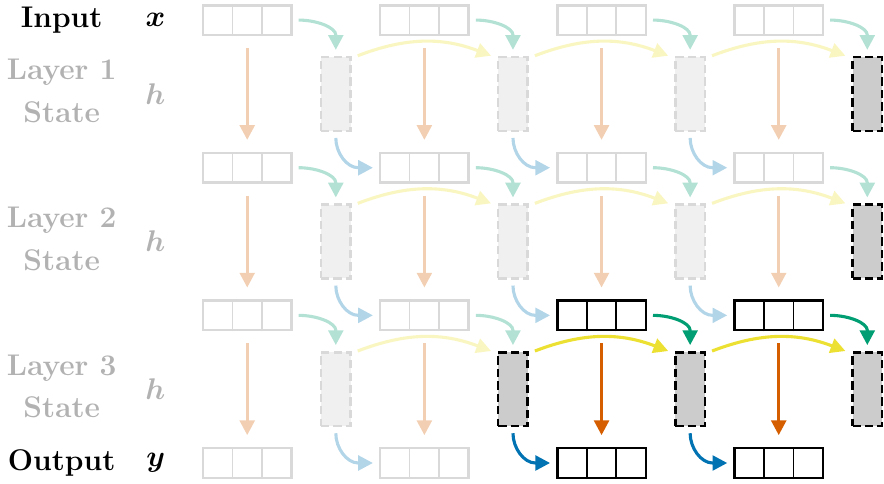}
    \caption{Fully vertical inference: parallelization is limited to a fixed vertical chunk size, but only the activations of the current chunk and one recurrent state per layer need to be stored.}
    \label{fig:cross-layer-inference:vertical}
  \end{subfigure}
  \caption{Final forward pass through the last model layer using fully horizontal (left) and vertically chunked (right) inference. Highlighted regions indicate activations retained in memory. A visualization of the full forward pass through all layers is provided in Appendix~\ref{app:cross-layer-inference}.}
  \label{fig:cross-layer-inference}
\end{figure*}

\noindent
First, we calculate the intra-chunk contributions to the outputs and the boundary latent states for each chunk independently, assuming a zero incoming chunk-boundary state ($\hb^{(c-1)} = \zerovec$). These operations are chunk-local matrix transformations:
\begin{gather}
  \label{eq:chunked-inference:intra-output}
  \chunked{\y}_{\text{intra}}^{(c)} = \bigl(\chunked{\matL}^{(c,c)} \circ (\chunked{\C}^{(c)}\, \chunked{\B}^{(c)^\top})\bigr) \chunked{\x}^{(c)} \\[0.5em]
  \label{eq:chunked-inference:intra-state}
  \hb_{\text{intra}}^{(c)} = \chunked{\B}^{(c)^\top} \diag{\chunked{\matL}^{(c,c)}_{Q,\cdot}} \chunked{\x}^{(c)}
\end{gather}

\paragraph{Inter-chunk State Propagation (Sequential)}\mbox{}

\noindent
To maintain global context, we must account for the state left by the previous chunk. We recurrently calculate the final chunk-boundary state $\hb^{(c)}$ by combining the final boundary state of the previous chunk with the current intra-chunk contribution. Given an initial state $\hb^{(0)} = \zerovec$,
\begin{equation}
  \label{eq:chunked-inference:inter-state}
  \hb^{(c)} = \A^\times_{(c Q):((c - 1) Q)} \hb^{(c-1)} + \hb_{\text{intra}}^{(c)}
\end{equation}

\paragraph{Final Output Adjustment (Parallelizable)}\mbox{}

\noindent
Finally, we calculate the correction needed for the output of each chunk based on the updated final state from the previous chunk, add it to our intra-chunk result, and assemble the final output. We define the base case $\chunked{\y}_{\text{inter}}^{(1)} = \zerovec$, since the first chunk has no preceding context. For $c > 1$,
\begin{equation}
  \label{eq:chunked-inference:inter-output}
  \chunked{\y}_{\text{inter}}^{(c)} = \diag{\chunked{\matL}^{(c,c-1)}_{\cdot,Q}} \, \chunked{\C}^{(c)} \, \hb^{(c-1)}
\end{equation}
\vspace{-\abovedisplayskip-\belowdisplayskip}
\begin{subequations}
  \begin{align}
    \label{eq:chunked-inference:intra-inter-output-combination}
    \chunked{\y}^{(c)} &= \chunked{\y}_{\text{intra}}^{(c)} + \chunked{\y}_{\text{inter}}^{(c)} \\
    \y_{(c-1) Q + i} &= \chunked{\y}^{(c)}_{i}
  \end{align}
\end{subequations}

The remaining sequential computations for the state propagation in Equation~\ref{eq:chunked-inference:inter-state} operate only on $K$ chunk-boundary states and consist of matrix-vector multiplications with low computational overhead. The majority of computations can be performed in parallel. Consequently, the overall computational cost scales as $\bigO{K \cdot Q^2} = \bigO{T \cdot Q}$, with the activation memory following the same scaling.

\endgroup

\subsubsection{Cross-Layer Inference Strategies}
\label{sec:recurrent-embedding-models:cross-layer-inference}
Modern embedding models consist of $L$ layers stacked sequentially, where the output of layer $l$ serves as input to layer $l+1$. The standard execution pattern is \emph{horizontal inference}: the model processes the entire sequence of length~$T$ through one layer at a time. Each Mamba2 layer can internally leverage the chunked parallelization described in Section~\ref{sec:recurrent-embedding-models:chunked-inference} as illustrated in Figure~\ref{fig:cross-layer-inference:horizontal}. While this horizontal inference strategy benefits from intra-layer parallelization, it requires storing full-sequence intermediate activations for the currently processed layer. Consequently, the resulting memory consumption of $\bigO{K \cdot Q^2} = \bigO{T \cdot Q}$ scales linearly with the input length~$T$.

To overcome this limitation, we propose chunked \emph{vertical inference}, which trades sequence-level parallelization for depth-wise recurrence. 
Specifically, we partition the input sequence into $K' = T / V$ vertical chunks of size $V$, where $V$ is a multiple of the intra-layer chunk size $Q$, implying $V \ge Q$. We process a single vertical chunk through all $L$ layers before advancing to the next chunk, as illustrated in Figure~\ref{fig:cross-layer-inference:vertical}. For each layer, the recurrent state is preserved across vertical chunks, analogous to the within-layer state recurrence across time steps described in Section~\ref{sec:recurrent-embedding-models:chunked-inference}. This creates a \emph{cross-layer recurrence} over vertical chunks in addition to the within-layer recurrence over horizontal chunks. The memory footprint of the model is now decoupled from the total sequence length $T$ for $T > V$, since we only need to store the activations of the current vertical chunk of length $V$ and the $L$ recurrent states, one for each layer. For input lengths $T \le V$, we can fall back to standard horizontal inference, which avoids the need to store latent states across layers and thus eliminates any memory overhead from vertical chunking for short sequences. A detailed step-by-step visualization of both strategies is provided in Appendix~\ref{app:cross-layer-inference}.

The resulting memory consumption is $\bigO{L + Q^2 \cdot \frac{V}{Q}} = \bigO{L + QV}$, where the first term accounts for the $L$ per-layer recurrent states and the second for the activations of the current vertical chunk, remaining constant in $T$ for $T > V$. While reducing the number of tokens processed simultaneously from $T$ to $V$ theoretically limits parallelization, modern hardware often reaches its compute-bound ceiling at comparatively small sequence lengths.
By choosing a vertical chunk size $V$ large enough to saturate hardware throughput but small enough to fit within memory constraints, we achieve the speed of horizontal inference with the scalability of recurrence. We validate the efficiency of our approach empirically in Section~\ref{sec:experiments:inference-evaluation}, and provide practical parameter selection guidelines in Appendix~\ref{app:parameter-selection}.

\section{Experiments}
\label{sec:experiments}
\subsection{Model Comparison}

\begin{table*}[t]
  \centering
  \begin{tabular}{l@{\hspace{2em}}*{2}{wc{(\widthof{\textbf{MTEB(Mult., v2)}}-3\tabcolsep)/2}}@{\hspace{2em}}*{2}{wc{(\widthof{\textbf{MTEB(eng, v2)}}-3\tabcolsep)/2}}@{\hspace{2em}}c}
\toprule
\textbf{Model} & \multicolumn{2}{@{}c@{\hspace{2em}}}{\textbf{MTEB(Mult., v2)}} & \multicolumn{2}{@{}c@{\hspace{2em}}}{\textbf{MTEB(eng, v2)}} & \textbf{LongEmbed} \\
Mean over $\rightarrow$ & Task & Type & Task & Type & Task \\
\midrule[\heavyrulewidth]
\modelqwen & 56.2 & 48.7 & 65.7 & 62.3 & 41.0 \\
\modelmamba & 55.2 & 47.9 & 64.3 & 60.9 & 40.8 \\
\midrule
\modelmistralversion{v0.1} & 58.6 & 50.5 & 67.3 & 63.3 & 44.7 \\
\modelcodestral & 59.4 & 51.9 & 65.2 & 61.8 & 44.5 \\
\bottomrule
\end{tabular}

  \caption{Evaluation of our fine-tuned transformer-based and recurrent embedding models on \multilingualmteb, \englishmteb, and \longembed. Results are aggregated as the mean over tasks and the mean over task types where applicable. Detailed per-task-type results for all benchmarks are provided in Appendix~\ref{app:evaluation-details}.}
  \label{tab:results-mteb-model-comparison}
\end{table*}

\begin{table}[t]
  \centering
  \begin{tabular}{l@{\hspace{2em}}*{2}{wc{(\widthof{\textbf{MTEB(eng, v1)}}-3\tabcolsep)/2}}}
\toprule
\textbf{Model} & \multicolumn{2}{@{}c}{\textbf{MTEB(eng, v1)}} \\
Mean over $\rightarrow$ & Task & Type \\
\midrule[\heavyrulewidth]
\multicolumn{3}{c}{\small \color{black!75} Results from \citet{wang-etal-2024-improving-text}} \\\addlinespace[0.2em]
\modelefivewithversion & 66.5 & 64.2 \\
\midrule
\multicolumn{3}{c}{\small \color{black!75} Results from \citet{springer-etal-2025-understanding}} \\\addlinespace[0.2em]
\modelqwen & 63.3 & 61.3 \\
\modelmistralversion{v0.1} & 65.5 & 63.2 \\
\midrule
\multicolumn{3}{c}{\small \color{black!75} Our results} \\
\addlinespace[-0.4em]\multicolumn{3}{c}{\tiny \color{black!75} \modelefive\ training recipe with data from \citet{springer-etal-2025-understanding}} \\\addlinespace[0.2em]
\modelqwen & 63.6 & 61.8 \\
\modelmistralversion{v0.1} & 65.7 & 63.5 \\
\bottomrule
\end{tabular}

  \caption{Evaluation of our fine-tuned transformer-based models on \englishmtebold, alongside scores reported by \citet{wang-etal-2024-improving-text} and \citet{springer-etal-2025-understanding}. Results are aggregated as the mean over tasks and the mean over task types where applicable. Detailed per-task-type results are provided in Appendix~\ref{app:evaluation-details}.}
  \label{tab:results-mteb-reproduction}
\end{table}

We demonstrate the effectiveness of recurrent embedding models by fine-tuning pretrained language models based on both Mamba2 and transformer architectures using the training procedure outlined in Section~\ref{sec:recurrent-embedding-models:training}. The resulting embedding models are evaluated on the original English Massive Text Embedding Benchmark \englishmtebold, its updated version \englishmteb, the Massive Multilingual Text Embedding Benchmark \multilingualmteb, and the long-context retrieval benchmark \longembed~\citep{muennighoff-etal-2023-mteb,enevoldsen_mmteb_2025,zhu-etal-2024-longembed}.

The \englishmteb\ benchmark includes 41 English datasets covering classification, clustering, pair classification, reranking, retrieval, semantic textual similarity (STS), and summarization tasks. The \multilingualmteb\ benchmark extends this to a multilingual setting encompassing more than 250 languages, and \longembed\ evaluates retrieval performance on long documents with sequences of up to 32,768 tokens in our setup.

We compare our fine-tuned transformer and recurrent models (Table~\ref{tab:results-mteb-model-comparison}) after validating our training procedure (Table~\ref{tab:results-mteb-reproduction}). First, we reproduce the results of \modelefive\ \citep{wang-etal-2024-improving-text} and the reproduction by \citet{springer-etal-2025-understanding} by fine-tuning pretrained checkpoints of \modelmistralversion{v0.1}\footnotemark\ \citep{jiang_mistral_2023}, the base model underlying \modelefive, and \modelqwen\ \citep{yang_qwen2_2024}. We train the models for one epoch on a combination of public datasets and the synthetic data generated using Llama~3.1~70B~\citep{grattafiori_llama_2024} and published by \citet{springer-etal-2025-understanding}, which replicates the synthetic data generation pipeline of \modelefive. Further details on the training data and hyperparameters can be found in Appendix~\ref{app:training-details}.

The results are shown in Table~\ref{tab:results-mteb-reproduction}.
Our fine-tuned \modelqwen\ improves upon the task mean reported by \citet{springer-etal-2025-understanding} by $0.3$ percentage points, while our model based on \modelmistralversion{v0.1} improves by $0.2$ percentage points, reducing the gap to the original \modelefive\ scores.
\footnotetext{
  We additionally fine-tune the more recent version \version{v0.3} of \modelmistral, which is used in the inference evaluation in Section~\ref{sec:experiments:inference-evaluation}. The differences between the versions and detailed per-task results for both versions are provided in Appendix~\ref{app:training-details} and Appendix~\ref{app:evaluation-details}, respectively.
}

For comparison, we fine-tune pretrained Mamba2-based language models of comparable size following the same training procedure, data, and hyperparameters where applicable. Concretely, we use the \modelmamba\ model published by \citet{dao_transformers_2024} and the code generation model \modelcodestral\footnote{\url{https://mistral.ai/news/codestral-mamba}}. 
We found that \modelmamba\ benefits from a considerably higher learning rate compared to the other models. Details on the Mamba2 base models and the learning rate selection for \modelmamba\ are provided in Appendix~\ref{app:training-details}.

\begin{figure*}[t]
  \includegraphics[width=\linewidth]{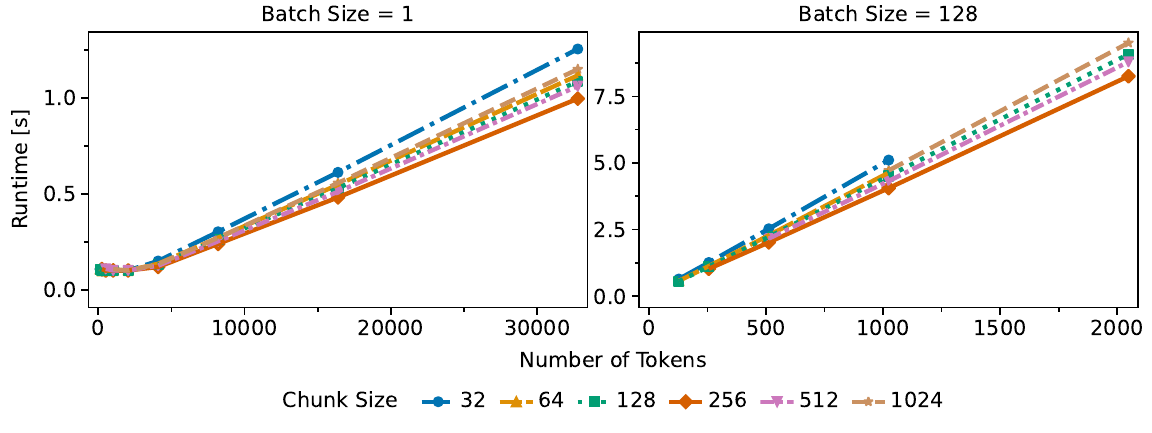}
  \caption{Runtime for \modelcodestral\ using fully horizontal chunked inference with varying chunk sizes for batch sizes 1 and 128.}
  \label{fig:inference-chunk-sizes}
\end{figure*}

In Table~\ref{tab:results-mteb-model-comparison}, we contrast our fine-tuned transformer and recurrent models under largely identical fine-tuning conditions across three benchmarks. The performance differences of recurrent and transformer-based models vary between benchmarks. On \multilingualmteb, \modelcodestral\ outperforms \modelmistralversion{v0.1}, achieving the highest mean task score among our fine-tuned models, while \modelmamba\ trails \modelqwen\ by a comparable margin. For \englishmteb, the Mamba2-based models lag behind their transformer-based counterparts by 1--2 percentage points. On \longembed, both model families achieve comparable performance, with differences remaining within $0.2$ percentage points for models of similar size. These results demonstrate that recurrent models are viable general-purpose text embedders, achieving competitive quality while offering favorable inference efficiency. The remaining quality gap on English tasks may be partially attributable to differences in pretraining scale and data composition between the recurrent and transformer base models, as \modelmamba\ was pretrained on significantly less data and \modelcodestral\ was specifically trained for code generation. We anticipate that further improvements are possible with more extensively pretrained base models.

\subsection{Inference Evaluation}
\label{sec:experiments:inference-evaluation}

\begin{figure*}[p]
  \includegraphics[width=\linewidth]{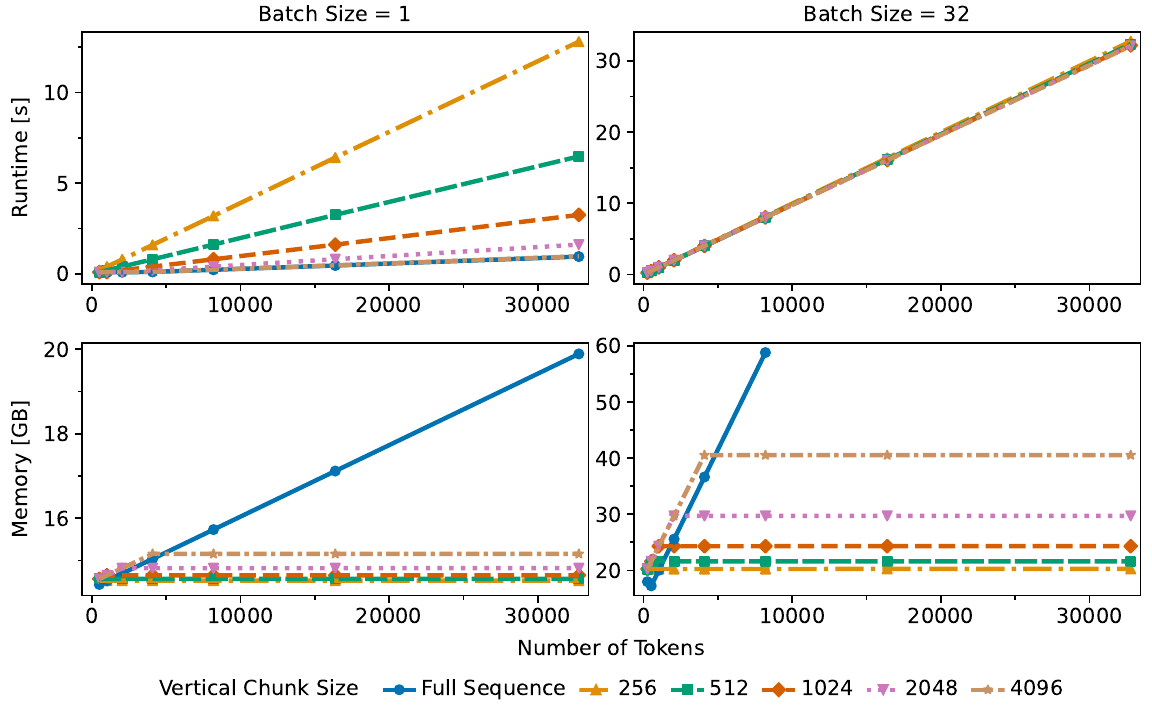}
  \caption{Runtime and memory usage for \modelcodestral\ using vertically chunked inference with a fixed intra-layer chunk size of 256 and varying vertical chunk sizes for batch sizes 1 and 32.}
  \label{fig:inference-vertical-chunk-sizes}
\end{figure*}

\begin{figure*}[p]
  \includegraphics[width=\linewidth]{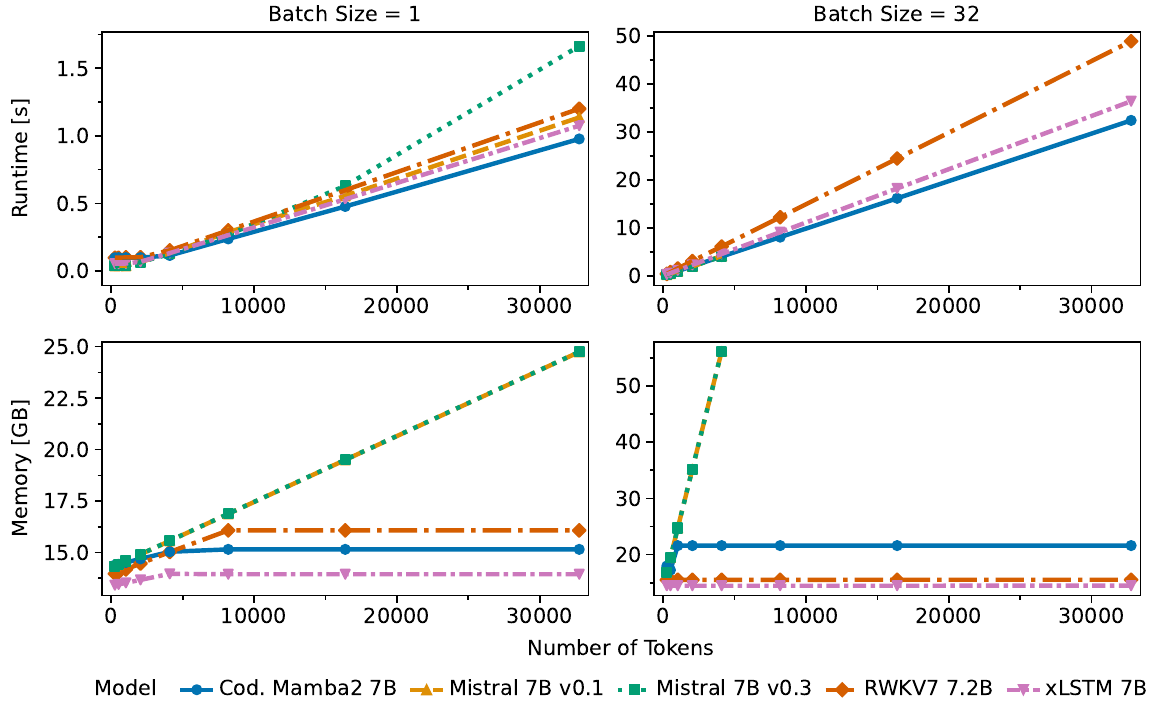}
  \caption{Runtime and memory usage for \modelcodestral, \modelrwkv, and \modelxlstm\ compared to \modelmistralversion{v0.1} and \version{v0.3}. \modelcodestral\ uses vertically chunked inference with an intra-layer chunk size of 256 and vertical chunk sizes of 4096 (batch size 1) and 512 (batch size 32). Latent states are stored across layers only for input lengths exceeding the vertical chunk size. \modelrwkv\ and \modelxlstm\ use the same strategy with chunk sizes 32 and 64 and vertical chunk sizes 64 and 128, respectively, for batch size 32. Both use a vertical chunk size of 4096 for batch size 1. Transformer-based models use FlashAttention-2.}
  \label{fig:inference-model-comparison}
\end{figure*}

To investigate the efficiency of recurrent embedding inference strategies, we measure runtime and memory consumption during embedding generation for varying input lengths and batch sizes on a single NVIDIA H100 with 80\,GB of memory. We evaluate \modelcodestral\ using both the fully horizontal and the vertically chunked inference strategies outlined in Section~\ref{sec:recurrent-embedding-models:inference}, with varying chunk sizes to explore the effect of different parallelization levels. For the model comparison, we additionally evaluate \modelrwkv\ and \modelxlstm\ using their respective optimal vertically chunked inference configurations. For comparison against transformer-based models, we measure the runtime and memory consumption of \modelmistralversion{v0.1} and \modelmistralversion{v0.3}. We repeat our experiments three times and report mean results to account for system load variability. Corresponding minimum and maximum values are reported in Appendix~\ref{app:parameter-robustness-portability}.

Figure~\ref{fig:inference-chunk-sizes} visualizes the effect of different chunk sizes on the runtime of \modelcodestral\ for single-sequence embedding generation and for batches of 128 sequences. We observe that a chunk size of $Q = 256$ provides the best performance across input lengths and batch sizes. Smaller chunk sizes reduce the parallelization potential, while larger chunk sizes increase the overhead of quadratic intra-chunk computations. We use this chunk size as the basis for further experiments.

With $Q$ fixed at $256$, Figure~\ref{fig:inference-vertical-chunk-sizes} shows that the single-sequence runtime approaches that of fully horizontal inference as the vertical chunk size ($V$) increases, reaching full convergence at $V = 4096$ in our setup. This indicates that parallelization is fully saturated at 4096 tokens, making the performance impact of cross-layer recurrence negligible. For a batch size of 32, the runtime remains close to that of fully horizontal inference across all vertical chunk sizes, showing only a minor increase for a vertical chunk size of $V = 256$. Parallelization saturates at smaller vertical chunk sizes when processing batches. 

A cross-architecture comparison of runtime and memory consumption against the transformer-based \modelmistralversion{v0.1} and \version{v0.3} models is shown in Figure~\ref{fig:inference-model-comparison}. For single-sequence embedding generation, \modelcodestral\ outperforms both transformer-based models in terms of runtime and memory usage for input lengths beyond 4096 tokens. The advantage over \modelmistralversion{v0.3} grows with sequence length due to the quadratic complexity of full attention, while \modelmistralversion{v0.1} does not exhibit this quadratic growth since it uses a sliding window of 4096 tokens. The runtime of all models is comparable for batches of 32 sequences, although both \modelmistral\ versions were only evaluated for input lengths up to 4096 tokens due to memory requirements exceeding the available accelerator memory. All three recurrent architectures share the same memory-scaling pattern: consumption grows linearly up to the vertical chunk size and then remains practically constant for longer sequences, in contrast to the linear growth of the transformer-based models. As a result, all recurrent models exhibit substantially lower memory requirements than transformers for long sequences.

\section{Conclusions}
We have demonstrated that recurrent language models are a viable and efficient foundation for text embedding generation. By applying standard contrastive fine-tuning, our Mamba2-based models achieve performance competitive with similarly sized transformers across the \englishmteb, \multilingualmteb, and \longembed\ benchmarks. Minor performance differences exist on individual benchmarks, with transformers slightly leading on English tasks and recurrent models showing advantages on multilingual tasks, while overall results remain consistently close.

Our primary technical contribution is a vertically chunked inference strategy applicable to recurrent architectures with a linear-time recurrence and a dual, parallelizable formulation. By introducing a cross-layer scheduling approach that processes data in fixed-size vertical blocks, we avoid storing full-sequence activations between layers, reducing memory requirements to a function of model depth rather than sequence length. Empirical validation on Mamba2, RWKV, and xLSTM models confirms consistent behavior across all three architectures: activation memory becomes effectively constant beyond the vertical chunk size, and parallelism saturates at relatively small vertical chunk sizes. In practice, the vertical chunk size can serve as a mechanism to enable inference under strict memory constraints while retaining close to fully parallel runtime, rather than as a hyperparameter requiring careful optimization.

Our comparisons reveal that while transformers and recurrent models perform similarly for short context lengths, the recurrent approach offers favorable runtime-memory trade-offs at moderate scale and becomes particularly advantageous for applications involving long contexts or batch processing, where the memory requirements of transformers become prohibitive. These findings establish recurrent architectures as a scalable and resource-efficient alternative for embedding generation. We anticipate that more extensively pretrained recurrent base models are a promising direction toward closing the remaining quality gap on English tasks, paving the way for a new class of long-context, resource-efficient embedding models.

\section*{Limitations}

\paragraph{Recurrent Architectures.} Our embedding quality evaluation focuses on Mamba2 as a representative recurrent architecture. While we extend the empirical inference efficiency comparison to \modelrwkv\ and \modelxlstm\ to validate the generalizability of vertically chunked inference, a thorough investigation of embedding quality for architectures beyond Mamba2 is left for future work.

\paragraph{Base Models.} The \modelmamba\ model, which serves as a foundation for our experiments, was pretrained on 700 billion tokens, an order of magnitude fewer than the \modelqwen\ transformer model, which was trained on 7 trillion tokens. For \modelcodestral, information on the exact pretraining data and duration is not publicly available, but as it is an instruction-tuned code generation model, its pretraining likely differs considerably from the transformer-based models evaluated in this work. This discrepancy may influence the embedding quality comparison and could explain the remaining small quality gaps. Fine-tuning more extensively pretrained recurrent base models under comparable pretraining conditions is a promising direction for future work.

\paragraph{Evaluation Scope.} While our experiments demonstrate the effectiveness of Mamba2-based embedding models across the \englishmteb, \multilingualmteb, and \longembed\ benchmarks, covering a wide range of downstream tasks, domains, and languages, other potential applications remain unevaluated. In particular, we do not account for the distribution of input sequence lengths in the benchmarks nor do we systematically vary input lengths for controlled evaluation. Further investigation is needed to assess the performance of recurrent embedding models across a wider range of input lengths and specialized domains.

\paragraph{Broader Impact.} Our work contributes to making text embedding generation more efficient, which can facilitate access to NLP technology in resource-constrained settings. However, embedding models can inherit biases present in their pretraining data. Users should be aware that embedding-based similarity judgments may reflect such biases and should supplement model outputs with expert verification where necessary.

\section*{Acknowledgments}
We thank the anonymous reviewers for their thoughtful and constructive feedback, which helped us strengthen both the experimental analysis and the presentation of this paper.

Emanuel Sallinger's work on this paper was funded by the Vienna Science and Technology Fund (WWTF), Grant IDs \texttt{10.47379/VRG18013}, \texttt{10.47379/ICT25032}, \texttt{10.47379/NXT22018}, \texttt{10.47379/ICT2201}, \texttt{10.47379/DCDH001}, and by the Austrian Science Fund (FWF) \texttt{10.55776/COE12}.

\bibliography{main}

\appendix

\onecolumn

\begingroup
\allowdisplaybreaks

\section{Chunked Inference Derivation}
\label{app:chunked-inference-derivation}

The calculation of $\chunked{\y}_{\text{intra}}^{(c)}$ in Equation~\ref{eq:chunked-inference:intra-output} corresponds to the diagonal block of the full SSM matrix $\M$ in the block decomposition of~\citet[Section~6]{dao_transformers_2024}, restricted to within-chunk interactions. Starting from Equation~\ref{eq:ssm-matrix:M} restricted to chunk $c$,
we rewrite the expression in chunked notation and factor it via Equation~\ref{eq:ssm-matrix-vectorized}:
\begin{subequations}
  \begin{align}
    \chunked{\y}_{\text{intra}}^{(c)} &=
    \begin{bmatrix}
      \C^\top_{(c-1)Q+1} \A^\times_{((c-1)Q+1):((c-1)Q+1)} \B_{(c-1)Q+1} & \cdots & 0 \\
      \vdots & \ddots & \vdots \\
      \C^\top_{cQ} \A^\times_{(cQ):((c-1)Q+1)} \B_{(c-1)Q+1} & \cdots & \C^\top_{cQ} \A^\times_{(cQ):(cQ)} \B_{cQ}
    \end{bmatrix}
    \chunked{\x}^{(c)} \\
    &=
    \begin{bmatrix}
      \chunked{\C}^{(c)^\top}_1 \chunked{\A}^{(c)^\times}_{1:1} \chunked{\B}^{(c)}_1 & \cdots & 0 \\
      \vdots & \ddots & \vdots \\
      \chunked{\C}^{(c)^\top}_Q \chunked{\A}^{(c)^\times}_{Q:1} \chunked{\B}^{(c)}_1 & \cdots & \chunked{\C}^{(c)^\top}_Q \chunked{\A}^{(c)^\times}_{Q:Q} \chunked{\B}^{(c)}_Q
    \end{bmatrix}
    \chunked{\x}^{(c)} \\
    &= \bigl(\chunked{\matL}^{(c,c)} \circ (\chunked{\C}^{(c)}\, \chunked{\B}^{(c)^\top})\bigr) \chunked{\x}^{(c)}
  \end{align}
\end{subequations}

\noindent
The intra-chunk state $\hb_{\text{intra}}^{(c)}$ in Equation~\ref{eq:chunked-inference:intra-state} computes the latent state at the end of chunk $c$ from inputs within that chunk alone, disregarding the state from preceding chunks. This corresponds to the right ($\B$) block factor of the low-rank off-diagonal blocks in the block decomposition of~\citet[Section~6]{dao_transformers_2024}, accumulating input contributions weighted by the row vector $\chunked{\matL}^{(c,c)}_{Q,\cdot}$:
\begin{subequations}
  \begin{align}
    \hb_{\text{intra}}^{(c)} &=
    \begin{bmatrix}
      \B^\top_{(c - 1) Q + 1} \A^\times_{(c Q):((c - 1) Q + 1)} \\
      \cdots \\
      \B^\top_{c Q} \A^\times_{(c Q):(c Q)}
    \end{bmatrix}^\top
    \chunked{\x}^{(c)} \\ 
    &= 
    {\underbrace{
      \begin{bmatrix}
        \chunked{\B}^{(c)^\top}_1 \chunked{\A}^{(c)^\times}_{Q:1} \\
        \cdots \\
        \chunked{\B}^{(c)^\top}_Q \chunked{\A}^{(c)^\times}_{Q:Q}
    \end{bmatrix}
    }_{\bigl(\chunked{\B}^{(c)^\top} \smalldiag{\chunked{\matL}^{(c,c)}_{Q,\cdot}}\bigr)^\top}}^\top
    \chunked{\x}^{(c)} \\
    &= \chunked{\B}^{(c)^\top} \diag{\chunked{\matL}^{(c,c)}_{Q,\cdot}} \chunked{\x}^{(c)}
  \end{align}
\end{subequations}

\noindent
The final chunk state $\hb^{(c)}$ in Equation~\ref{eq:chunked-inference:inter-state} accumulates contributions from all chunks up to and including chunk $c$. This can be expressed as a recurrence over chunk states, where the center ($\A$) block factor propagates the global state of the previous chunk $\hb^{(c-1)}$ forward through the transition matrices:
\begin{subequations}
  \begin{align}
    \hb^{(c)} &= \sum_{k=1}^{c} \A^\times_{(c Q):(k Q)}
    \begin{bmatrix}
        \B^\top_{(k - 1) Q + 1} \A^\times_{(k Q):((k - 1) Q + 1)} \\
        \cdots \\
        \B^\top_{k Q} \A^\times_{(k Q):(k Q)}
    \end{bmatrix}^\top
    \chunked{\x}^{(k)} \\
    &= \sum_{k=1}^{c} \A^\times_{(c Q):(k Q)}
    \underbrace{
      \begin{bmatrix}
          \chunked{\B}^{(k)^\top}_1 \chunked{\A}^{(k)^\times}_{Q:1} \\
          \cdots \\
          \chunked{\B}^{(k)^\top}_Q \chunked{\A}^{(k)^\times}_{Q:Q}
      \end{bmatrix}^\top
      \chunked{\x}^{(k)}
    }_{\hb_{\text{intra}}^{(k)}} \\
    &= \sum_{k=1}^{c} \A^\times_{(c Q):(k Q)} \hb_{\text{intra}}^{(k)} \\
    &= \underbrace{\A^\times_{(c Q):(c Q)}}_{\I} \hb_{\text{intra}}^{(c)} + \sum_{k=1}^{c-1} \A^\times_{(c Q):(k Q)} \hb_{\text{intra}}^{(k)} \\
    &= \hb_{\text{intra}}^{(c)} + \A^\times_{(c Q):((c-1) Q)} \underbrace{\sum_{k=1}^{c-1} \A^\times_{((c-1) Q):(k Q)} \hb_{\text{intra}}^{(k)}}_{\hb^{(c-1)}} \\
    &= \hb_{\text{intra}}^{(c)} + \A^\times_{(c Q):((c - 1) Q)} \hb^{(c-1)}
  \end{align}
\end{subequations}

\noindent
The inter-chunk output correction $\chunked{\y}_{\text{inter}}^{(c)}$ in Equation~\ref{eq:chunked-inference:inter-output} captures the contribution of all preceding chunks to the output of chunk $c$. 
We start from the off-diagonal blocks of $\M$ (Equation~\ref{eq:ssm-matrix:y}). These factor into left ($\C$), center ($\A$), and right ($\B$) block factors in the block decomposition of~\citet[Section~6]{dao_transformers_2024}. The derivation then collapses the right factor and input into the accumulated global state $\hb^{(c-1)}$, leaving the left factor applied to $\hb^{(c-1)}$:
\begin{subequations}
  \begin{align}
    \chunked{\y}_{\text{inter}}^{(c)} &= \sum_{k=1}^{c - 1}
    \begin{bmatrix}
        \C^\top_{(c - 1) Q + 1} \A^\times_{((c - 1) Q + 1):((c - 1) Q)} \\
        \cdots \\
        \C^\top_{c Q} \A^\times_{(c Q):((c - 1) Q)}
    \end{bmatrix}
    \A^\times_{((c-1) Q):(k Q)}
    \begin{bmatrix}
        \B^\top_{(k - 1) Q + 1} \A^\times_{(k Q):((k - 1) Q + 1)} \\
        \cdots \\
        \B^\top_{k Q} \A^\times_{(k Q):(k Q)}
    \end{bmatrix}^\top
    \chunked{\x}^{(k)} \\
    &= \sum_{k=1}^{c-1}
    \begin{bmatrix}
      \chunked{\C}^{(c)^\top}_1 \chunked{\matL}^{(c,c-1)}_{1,Q} \\
        \cdots \\
      \chunked{\C}^{(c)^\top}_Q \chunked{\matL}^{(c,c-1)}_{Q,Q}
    \end{bmatrix}
    \A^\times_{((c-1) Q):(k Q)}
    \begin{bmatrix}
        \chunked{\B}^{(k)^\top}_1 \chunked{\A}^{(k)^\times}_{Q:1} \\
        \cdots \\
        \chunked{\B}^{(k)^\top}_Q \chunked{\A}^{(k)^\times}_{Q:Q}
    \end{bmatrix}^\top
    \chunked{\x}^{(k)} \\
    &= \underbrace{
      \begin{bmatrix}
        \chunked{\C}^{(c)^\top}_1 \chunked{\matL}^{(c,c-1)}_{1,Q} \\
        \cdots \\
        \chunked{\C}^{(c)^\top}_Q \chunked{\matL}^{(c,c-1)}_{Q,Q}
      \end{bmatrix}
    }_{\bigl(\chunked{\C}^{(c)^\top} \smalldiag{\chunked{\matL}^{(c,c-1)}_{\cdot,Q}}\bigr)^\top}
    \underbrace{
      \sum_{k=1}^{c-1}
      \A^\times_{((c-1) Q):(k Q)}
      \begin{bmatrix}
          \chunked{\B}^{(k)^\top}_1 \chunked{\A}^{(k)^\times}_{Q:1} \\
          \cdots \\
          \chunked{\B}^{(k)^\top}_Q \chunked{\A}^{(k)^\times}_{Q:Q}
      \end{bmatrix}^\top
      \chunked{\x}^{(k)}
    }_{\hb^{(c-1)}} \\
    &= \Bigl(\chunked{\C}^{(c)^\top} \diag{\chunked{\matL}^{(c,c-1)}_{\cdot,Q}}\Bigr)^\top \hb^{(c-1)} \\
    &= \diag{\chunked{\matL}^{(c,c-1)}_{\cdot,Q}} \, \chunked{\C}^{(c)} \, \hb^{(c-1)}
  \end{align}
\end{subequations}

\noindent
Finally, Equation~\ref{eq:chunked-inference:intra-inter-output-combination} shows that $\chunked{\y}^{(c)}$ decomposes additively into the within-chunk contribution (diagonal block) and the cross-chunk correction (off-diagonal blocks), corresponding to the block decomposition of $\M$ in~\citet[Section~6]{dao_transformers_2024}. Starting from $\y = \M \x$ (Equation~\ref{eq:ssm-matrix:y}) and restricting to chunk $c$:
\begin{subequations}
  \begin{align}
    \chunked{\y}^{(c)} & =
    \underbrace{
      \begin{bmatrix}
        \C^\top_{(c-1)Q+1} \A^\times_{((c-1)Q+1):((c-1)Q+1)} \B_{(c-1)Q+1} & \cdots & 0 \\
        \vdots & \ddots & \vdots \\
        \C^\top_{cQ} \A^\times_{(cQ):((c-1)Q+1)} \B_{(c-1)Q+1} & \cdots & \C^\top_{cQ} \A^\times_{(cQ):(cQ)} \B_{cQ}
      \end{bmatrix}
      \chunked{\x}^{(c)}
    }_{\chunked{\y}_{\text{intra}}^{(c)}} \nonumber \\
    & \quad + 
    \underbrace{
      \sum_{k=1}^{c - 1}
      \begin{bmatrix}
          \C^\top_{(c - 1) Q + 1} \A^\times_{((c-1)Q+1):((c - 1) Q)} \\
          \cdots \\
          \C^\top_{c Q} \A^\times_{(c Q):((c - 1) Q)}
      \end{bmatrix}
      \A^\times_{((c-1) Q):(k Q)}
      \begin{bmatrix}
          \B^\top_{(k - 1) Q + 1} \A^\times_{(k Q):((k - 1) Q + 1)} \\
          \cdots \\
          \B^\top_{k Q} \A^\times_{(k Q):(k Q)}
      \end{bmatrix}^\top
      \chunked{\x}^{(k)}
    }_{\chunked{\y}_{\text{inter}}^{(c)}} \\
    &= \chunked{\y}_{\text{intra}}^{(c)} + \chunked{\y}_{\text{inter}}^{(c)}
  \end{align}
\end{subequations}

\noindent
The diagonal block equals the within-chunk SSM output with zero initial state (Equation~\ref{eq:chunked-inference:intra-output}). The off-diagonal blocks factor into the $\C$ block factor applied to the accumulated global state (Equation~\ref{eq:chunked-inference:inter-output}).

\endgroup

\twocolumn

\section{Recurrent Architectures beyond Mamba2}
\label{app:recurrent-architectures}

Beyond structured state space models like Mamba2, a variety of recurrent architectures exhibit a dual formulation with both parallel and recurrent computation modes. The RWKV model family~\citep{peng-etal-2023-rwkv,peng_rwkv_2025} replaces self-attention with a time-decayed weighted key-value (WKV) operator which computes outputs either via a parallel scan over all time steps or through a lightweight per-step recurrence. Similarly, xLSTM~\citep{beck_xlstm_2024} introduces extended Long Short-Term Memory (LSTM) blocks with exponential gating and structured memory updates designed to be associative over time, enabling implementation both as parallel prefix operations across the sequence and as standard recurrent updates. These properties enable the combination of parallel intra-chunk computations with inter-chunk recurrence, analogous to the chunked inference strategy we present for Mamba2 in Section~\ref{sec:recurrent-embedding-models:chunked-inference}.

\section{Cross-Layer Inference}
\label{app:cross-layer-inference}

\begin{figure*}[t]
  \begin{subfigure}[b]{0.48\linewidth}
    \includegraphics[width=\linewidth]{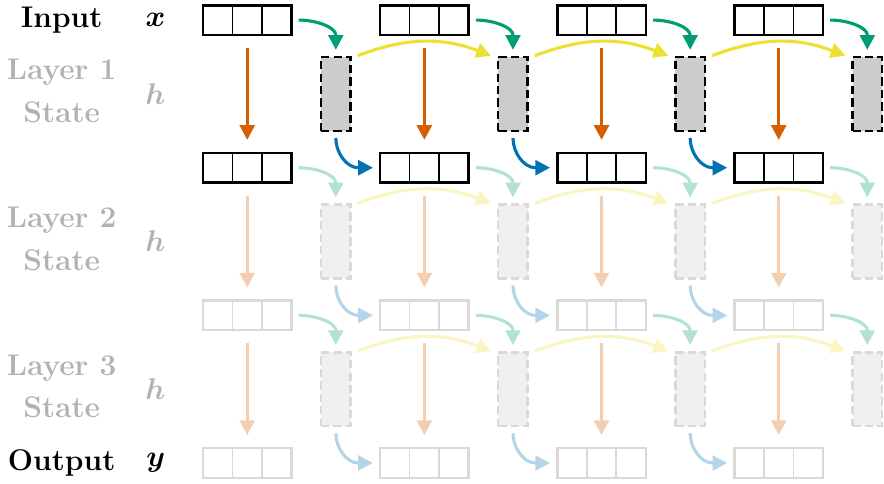}
    \vspace{-0.75em}
    \caption{Layer 1}
    \label{fig:horizontal-cross-layer-inference:layer-1}
  \end{subfigure}
  \hfill
  \begin{subfigure}[b]{0.48\linewidth}
    \includegraphics[width=\linewidth]{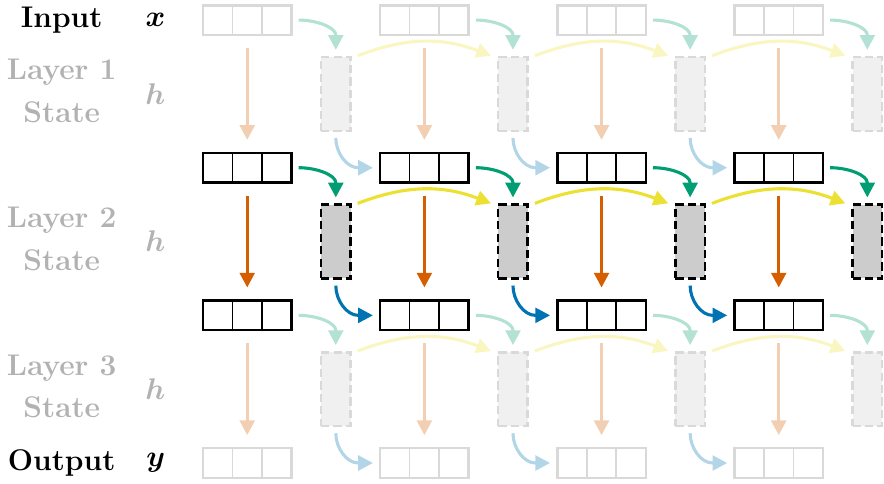}
    \vspace{-0.75em}
    \caption{Layer 2}
    \label{fig:horizontal-cross-layer-inference:layer-2}
  \end{subfigure}
  \vspace{1.25em}
  \\
  \centering
  \begin{subfigure}[b]{0.48\linewidth}
    \includegraphics[width=\linewidth]{figures/chunked-inference-horizontal-layer3}
    \vspace{-0.75em}
    \caption{Layer 3}
    \label{fig:horizontal-cross-layer-inference:layer-3}
  \end{subfigure}
  \caption{Step-by-step visualization of fully horizontal inference. Each diagram shows the state after processing one layer. All chunks are computed in parallel within each layer before advancing to the next, resulting in the accumulation of activations over the entire sequence.}
  \label{fig:horizontal-cross-layer-inference}
\end{figure*}

\begin{figure*}[t!]
  \begin{subfigure}[b]{0.48\linewidth}
    \includegraphics[width=\linewidth]{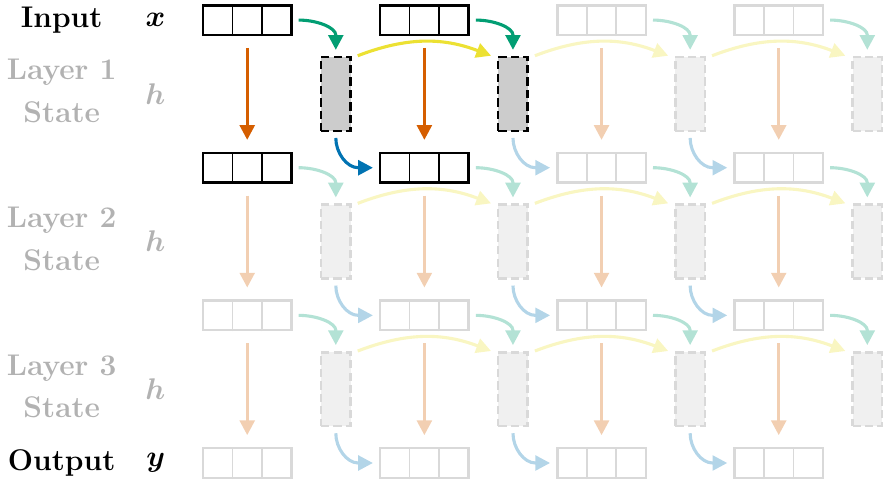}
    \vspace{-0.75em}
    \caption{Vertical Chunk 1, Layer 1}
    \label{fig:vertical-cross-layer-inference:layer-1-chunk-1}
  \end{subfigure}
  \hfill
  \begin{subfigure}[b]{0.48\linewidth}
    \includegraphics[width=\linewidth]{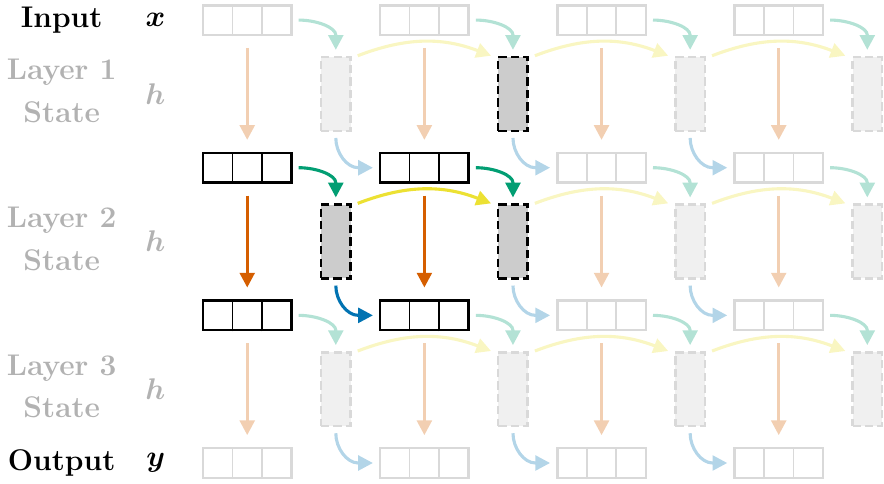}
    \vspace{-0.75em}
    \caption{Vertical Chunk 1, Layer 2}
    \label{fig:vertical-cross-layer-inference:layer-2-chunk-1}
  \end{subfigure}
  \vspace{1.25em}
  \\
  \begin{subfigure}[b]{0.48\linewidth}
    \includegraphics[width=\linewidth]{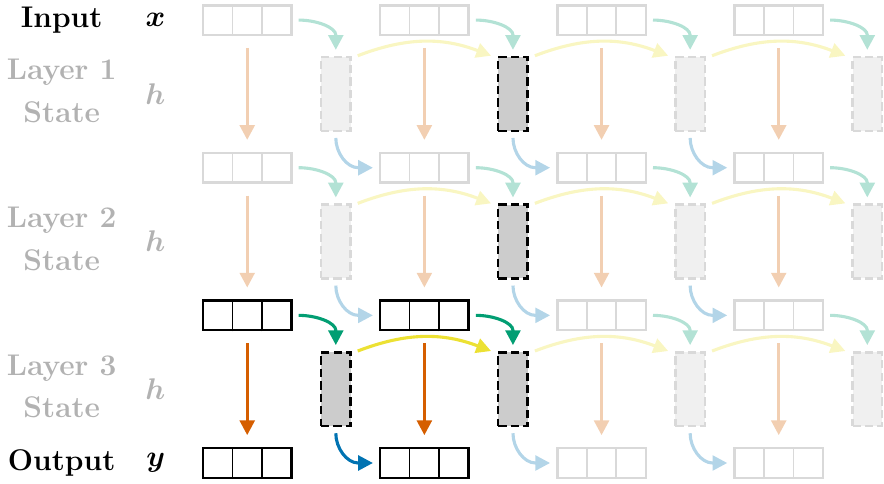}
    \vspace{-0.75em}
    \caption{Vertical Chunk 1, Layer 3}
    \label{fig:vertical-cross-layer-inference:layer-3-chunk-1}
  \end{subfigure}
  \hfill
  \begin{subfigure}[b]{0.48\linewidth}
    \includegraphics[width=\linewidth]{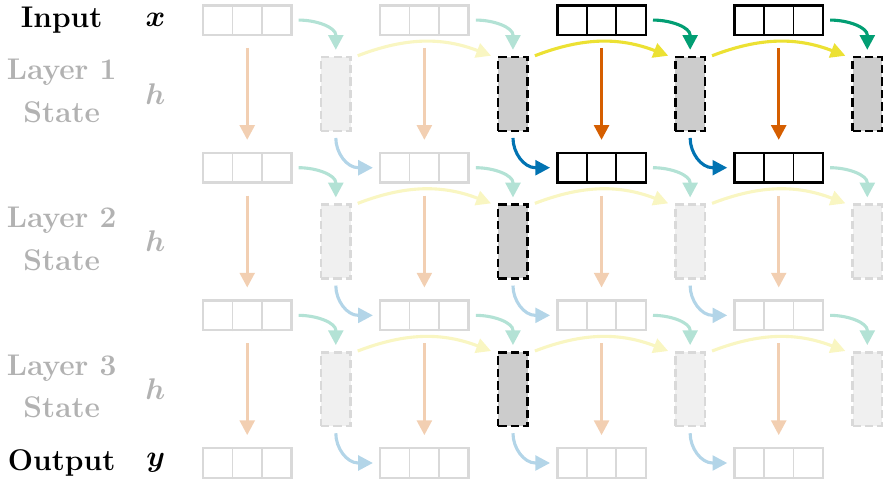}
    \vspace{-0.75em}
    \caption{Vertical Chunk 2, Layer 1}
    \label{fig:vertical-cross-layer-inference:layer-1-chunk-2}
  \end{subfigure}
  \vspace{1.25em}
  \\
  \begin{subfigure}[b]{0.48\linewidth}
    \includegraphics[width=\linewidth]{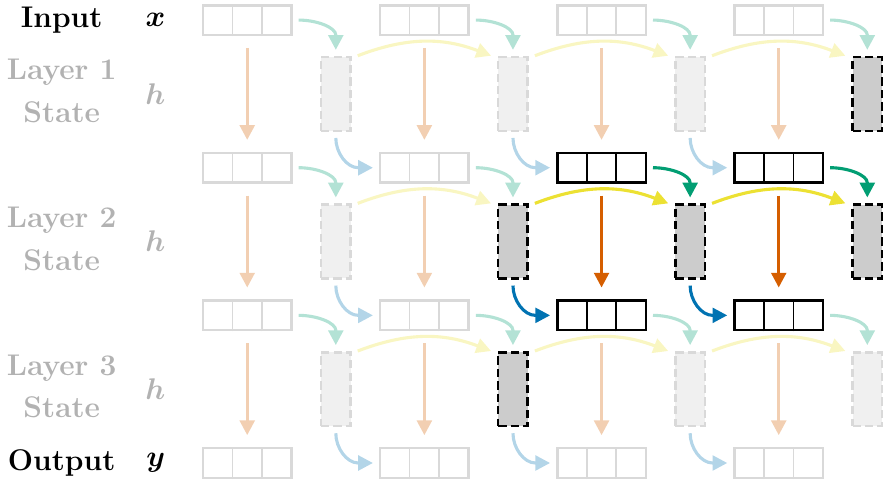}
    \vspace{-0.75em}
    \caption{Vertical Chunk 2, Layer 2}
    \label{fig:vertical-cross-layer-inference:layer-2-chunk-2}
  \end{subfigure}
  \hfill
  \begin{subfigure}[b]{0.48\linewidth}
    \includegraphics[width=\linewidth]{figures/chunked-inference-vertical-chunk2-layer3}
    \vspace{-0.75em}
    \caption{Vertical Chunk 2, Layer 3}
    \label{fig:vertical-cross-layer-inference:layer-3-chunk-2}
  \end{subfigure}
  \caption{Step-by-step visualization of vertically chunked inference. Each diagram shows one computation step, progressing through all layers for the first vertical chunk before advancing to the second vertical chunk. Only the activations of the current vertical chunk and one state per layer must be retained in memory.}
  \label{fig:vertical-cross-layer-inference}
\end{figure*}

In Section~\ref{sec:recurrent-embedding-models:cross-layer-inference}, we introduced two cross-layer inference strategies: fully horizontal inference and vertically chunked inference. Whereas the main text highlights the final step of these strategies to illustrate their memory footprint, this section details how each strategy processes data layer by layer. Figures~\ref{fig:cross-layer-inference:horizontal} and~\ref{fig:cross-layer-inference:vertical} show the final forward pass through the last model layer, while Figures~\ref{fig:horizontal-cross-layer-inference} and~\ref{fig:vertical-cross-layer-inference} illustrate the complete progression through all layers. They visualize how horizontal inference processes the entire sequence through each layer sequentially, accumulating intermediate activations for all tokens. In contrast, our vertical approach processes fixed-size chunks through all layers before advancing, maintaining only the current chunk's activations and per-layer recurrent states.

\section{Evaluation Details}
\label{app:evaluation-details}

\subsection{Embedding Quality}

\begin{table}[t]
  \centering
  \begin{tabular}{ll}
\toprule
Task Type & Abbreviation \\
\midrule[\heavyrulewidth]
Bitext Mining & Bit.M. \\
Classification & Cls. \\
Clustering & Clust. \\
Instruction Reranking & I.Rera. \\
Multilabel Classification & M.Cls. \\
Pair Classification & P.Cls. \\
Reranking & Rera. \\
Retrieval & Retr. \\
STS & STS \\
Summarization & Sum. \\
\bottomrule
\end{tabular}

  \caption{\mteb\ task type abbreviations.}  \label{tab:mteb-task-type-abbreviations}
\end{table}

\begingroup
\setlength{\tabcolsep}{2pt}
\begin{table*}[t]
  \centering
  \begin{tabular}{l@{\hspace{2em}}ccccccccc@{\hspace{2em}}cc}
\toprule
\textbf{Model} & \textbf{Bit.M.} & \textbf{Cls.} & \textbf{Clust.} & \textbf{I.Rera.} & \textbf{M.Cls.} & \textbf{P.Cls.} & \textbf{Rera.} & \textbf{Retr.} & \textbf{STS} & \multicolumn{2}{c}{\textbf{Mean}}\\
\# of datasets $\rightarrow$ & 13 & 43 & 16 & 3 & 5 & 11 & 6 & 18 & 16 & Task & Type \\
\midrule[\heavyrulewidth]
&\multicolumn{9}{c@{\hspace{2em}}}{\small \color{black!75} Results from \citet{wang-etal-2024-improving-text}}&& \\[0.2em]
\modelefivewithversion & 70.6 & 60.3 & 50.6 & -0.6 & 22.2 & 81.1 & 63.8 & 55.8 & 74.0 & 60.2 & 53.1 \\
\midrule
&\multicolumn{9}{c@{\hspace{2em}}}{\small \color{black!75} Fine-tuned Transformers (ours)}&& \\[0.2em]
\modelqwen & 61.8 & 57.6 & 46.9 & -2.2 & 20.0 & 78.2 & 53.7 & 50.7 & 71.4 & 56.2 & 48.7 \\
\modelmistralversion{v0.1} & 68.6 & 60.2 & 46.9 & -4.7 & 21.2 & 79.9 & 55.3 & 54.4 & 72.4 & 58.6 & 50.5 \\
\modelmistralversion{v0.3} & 68.7 & 60.2 & 48.0 & -4.4 & 21.7 & 80.1 & 55.5 & 55.1 & 72.7 & 58.9 & 50.8 \\
\midrule
&\multicolumn{9}{c@{\hspace{2em}}}{\small \color{black!75} Fine-tuned Recurrent Models (ours)}&& \\[0.2em]
\modelmamba & 61.1 & 55.7 & 45.2 & -5.6 & 19.5 & 77.6 & 55.4 & 50.9 & 70.9 & 55.2 & 47.9 \\
\modelcodestralshort & 70.6 & 60.0 & 47.0 & -3.2 & 21.8 & 81.0 & 60.1 & 56.3 & 73.3 & 59.4 & 51.9 \\
\bottomrule
\end{tabular}

  \caption{Detailed results of the evaluated models on the \multilingualmteb\ benchmark. Abbreviations for task types are listed in Table~\ref{tab:mteb-task-type-abbreviations}.}
  \label{tab:results-mteb-multilingual-v2}
\end{table*}
\endgroup

\begin{table*}[t]
  \centering
  \begin{tabular}{l@{\hspace{2em}}ccccccc@{\hspace{2em}}cc}
\toprule
\textbf{Model} & \textbf{Cls.} & \textbf{Clust.} & \textbf{P.Cls.} & \textbf{Rera.} & \textbf{Retr.} & \textbf{STS} & \textbf{Sum.} & \multicolumn{2}{c}{\textbf{Mean}}\\
\# of datasets $\rightarrow$ & 8 & 8 & 3 & 2 & 10 & 9 & 1 & Task & Type \\
\midrule[\heavyrulewidth]
&\multicolumn{7}{c@{\hspace{2em}}}{\small \color{black!75} Results from \citet{wang-etal-2024-improving-text}}&& \\[0.2em]
\modelefivewithversion & 79.9 & 51.4 & 88.4 & 49.8 & 57.6 & 84.3 & 36.6 & 68.0 & 64.0 \\
\midrule
&\multicolumn{7}{c@{\hspace{2em}}}{\small \color{black!75} Fine-tuned Transformers (ours)}&& \\[0.2em]
\modelqwen & 78.1 & 49.7 & 85.2 & 48.4 & 54.1 & 82.2 & 38.3 & 65.7 & 62.3 \\
\modelmistralversion{v0.1} & 80.7 & 50.9 & 87.2 & 49.3 & 56.8 & 82.5 & 35.8 & 67.3 & 63.3 \\
\modelmistralversion{v0.3} & 79.9 & 52.0 & 87.1 & 49.3 & 56.4 & 82.4 & 37.1 & 67.3 & 63.5 \\
\midrule
&\multicolumn{7}{c@{\hspace{2em}}}{\small \color{black!75} Fine-tuned Recurrent Models (ours)}&& \\[0.2em]
\modelmamba & 76.8 & 48.6 & 85.0 & 48.3 & 50.7 & 81.9 & 35.0 & 64.3 & 60.9 \\
\modelcodestralshort & 79.5 & 50.5 & 86.2 & 49.3 & 50.3 & 81.7 & 35.2 & 65.2 & 61.8 \\
\bottomrule
\end{tabular}

  \caption{Detailed results of the evaluated models on the \englishmteb\ benchmark. Abbreviations for task types are listed in Table~\ref{tab:mteb-task-type-abbreviations}.}
  \label{tab:results-mteb-eng-v2}
\end{table*}

\begin{table*}[t]
  \centering
  \begin{tabular}{l@{\hspace{2em}}cccccc@{\hspace{2em}}c}
\toprule
\textbf{Model} & \textbf{NQA} & \textbf{Needle} & \textbf{Passkey} & \textbf{QMSum} & \textbf{SSFD} & \textbf{WikimQA} & \textbf{Mean}\\
\midrule[\heavyrulewidth]
&\multicolumn{6}{c@{\hspace{2em}}}{\small \color{black!75} Results from \citet{wang-etal-2024-improving-text}}& \\[0.2em]
\modelefivewithversion & 37.2 & 31.5 & 30.8 & 28.6 & 75.1 & 58.7 & 43.7 \\
\midrule
&\multicolumn{6}{c@{\hspace{2em}}}{\small \color{black!75} Fine-tuned Transformers (ours)}& \\[0.2em]
\modelqwen & 27.4 & 30.5 & 34.0 & 30.8 & 75.4 & 48.2 & 41.0 \\
\modelmistralversion{v0.1} & 40.8 & 28.5 & 37.8 & 31.9 & 79.8 & 49.6 & 44.7 \\
\modelmistralversion{v0.3} & 43.2 & 27.8 & 36.5 & 32.0 & 79.8 & 49.7 & 44.8 \\
\midrule
&\multicolumn{6}{c@{\hspace{2em}}}{\small \color{black!75} Fine-tuned Recurrent Models (ours)}& \\[0.2em]
\modelmamba & 24.2 & 32.2 & 37.8 & 29.9 & 72.9 & 47.9 & 40.8 \\
\modelcodestralshort & 35.7 & 31.0 & 38.8 & 30.2 & 75.0 & 56.5 & 44.5 \\
\bottomrule
\end{tabular}

  \caption{Detailed results of the evaluated models on the \longembed\ benchmark.}
  \label{tab:results-longembed}
\end{table*}

\begin{table*}[t]
  \centering
  \begin{tabular}{l@{\hspace{2em}}ccccccc@{\hspace{2em}}cc}
\toprule
\textbf{Model} & \textbf{Cls.} & \textbf{Clust.} & \textbf{P.Cls.} & \textbf{Rera.} & \textbf{Retr.} & \textbf{STS} & \textbf{Sum.} & \multicolumn{2}{c}{\textbf{Mean}}\\
\# of datasets $\rightarrow$ & 12 & 11 & 3 & 4 & 15 & 10 & 1 & Task & Type \\
\midrule[\heavyrulewidth]
&\multicolumn{7}{c@{\hspace{2em}}}{\small \color{black!75} Results from \citet{wang-etal-2024-improving-text}}&& \\[0.2em]
\modelefivewithversion & 77.4 & 50.3 & 88.4 & 60.2 & 57.1 & 84.7 & 31.5 & 66.5 & 64.2 \\
\midrule
&\multicolumn{7}{c@{\hspace{2em}}}{\small \color{black!75} Results from \citet{springer-etal-2025-understanding}}&& \\[0.2em]
\modelqwen & 76.5 & 47.2 & 86.9 & 55.0 & 54.4 & 79.3 & 29.7 & 63.3 & 61.3 \\
\modelmistralversion{v0.1} & 77.9 & 49.1 & 87.4 & 57.6 & 57.1 & 82.9 & 30.1 & 65.5 & 63.2 \\
\modelmistralversion{v0.2} & 78.3 & 50.5 & 88.2 & 60.0 & 58.2 & 85.7 & 31.3 & 66.9 & 64.6 \\
\midrule
&\multicolumn{7}{c@{\hspace{2em}}}{\small \color{black!75} Fine-tuned Transformers (ours)}&& \\[0.2em]
\modelqwen & 75.5 & 46.9 & 85.2 & 58.2 & 52.8 & 82.7 & 31.5 & 63.6 & 61.8 \\
\modelmistralversion{v0.1} & 77.6 & 49.5 & 87.2 & 59.9 & 56.3 & 82.9 & 31.0 & 65.7 & 63.5 \\
\modelmistralversion{v0.3} & 76.9 & 49.4 & 87.1 & 60.0 & 55.8 & 83.0 & 30.9 & 65.5 & 63.3 \\
\midrule
&\multicolumn{7}{c@{\hspace{2em}}}{\small \color{black!75} Fine-tuned Recurrent Models (ours)}&& \\[0.2em]
\modelmamba & 73.8 & 46.2 & 85.0 & 58.0 & 49.0 & 82.4 & 30.6 & 61.9 & 60.7 \\
\modelcodestralshort & 76.9 & 47.3 & 86.2 & 60.0 & 49.5 & 82.2 & 29.8 & 63.2 & 61.7 \\
\bottomrule
\end{tabular}

  \caption{Detailed results of the evaluated models on the \englishmtebold\ benchmark. Abbreviations for task types are listed in Table~\ref{tab:mteb-task-type-abbreviations}.}
  \label{tab:results-mteb-eng-v1}
\end{table*}

For our results in Section~\ref{sec:experiments}, we evaluate all models on the \englishmteb, \multilingualmteb, and \longembed\ benchmarks with \texttt{bfloat16} precision. For \englishmteb\ and \multilingualmteb, we limit the maximum sequence length to 512 following \citet{wang-etal-2024-improving-text} and \citet{springer-etal-2025-understanding} to ensure comparability. For \longembed, we evaluate all tasks with sequence lengths of up to 32,768 tokens, corresponding to the maximum context length of the models based on \modelmistral. The evaluation instructions are provided in Appendix~\ref{app:instructions-evaluation}.

Table~\ref{tab:mteb-task-type-abbreviations} defines the abbreviations used in 
Tables~\ref{tab:results-mteb-multilingual-v2}, \ref{tab:results-mteb-eng-v2}, \ref{tab:results-longembed}, and~\ref{tab:results-mteb-eng-v1}, presenting the detailed per-task and per-task-type results for the \multilingualmteb, \englishmteb, \longembed, and \englishmtebold\ benchmarks, respectively. These results provide a comprehensive breakdown of model performance, showcasing task-type-specific strengths and weaknesses.

\subsection{Analysis by Task Type}

The per-task-type results for \modelcodestral\ and \modelmistralversion{v0.1} are distinct for the English and multilingual benchmarks. 
\begin{itemize}
  \item \multilingualmteb, Table~\ref{tab:results-mteb-multilingual-v2}: \modelcodestral\ leads in bitext mining by $2.0$ points, reranking by $4.8$ points, and retrieval by $1.9$ points, closely matching or exceeding the transformer baseline across all task types.
  \item \englishmteb, Table~\ref{tab:results-mteb-eng-v2}: The pattern reverses for the English benchmark. \modelcodestral\ trails most notably on retrieval by $6.5$ points, which accounts for the majority of the overall score difference. The remaining English task types are close, with gaps staying approximately within 1--2 points on classification, pair classification, clustering, and STS.
  \item \longembed, Table~\ref{tab:results-longembed}: Performance is similar across all tasks, with \modelcodestral\ leading on one half of the tasks and \modelmistralversion{v0.1} leading on the other half.
\end{itemize}
In summary, the recurrent model outperforms or matches transformers on multilingual tasks, while English retrieval is the primary area where a notable quality gap remains. All other task types show near parity across all three benchmarks.

\section{Parameter Robustness and Portability}
\label{app:parameter-robustness-portability}

\begin{figure*}[ht]
  \includegraphics[width=\linewidth]{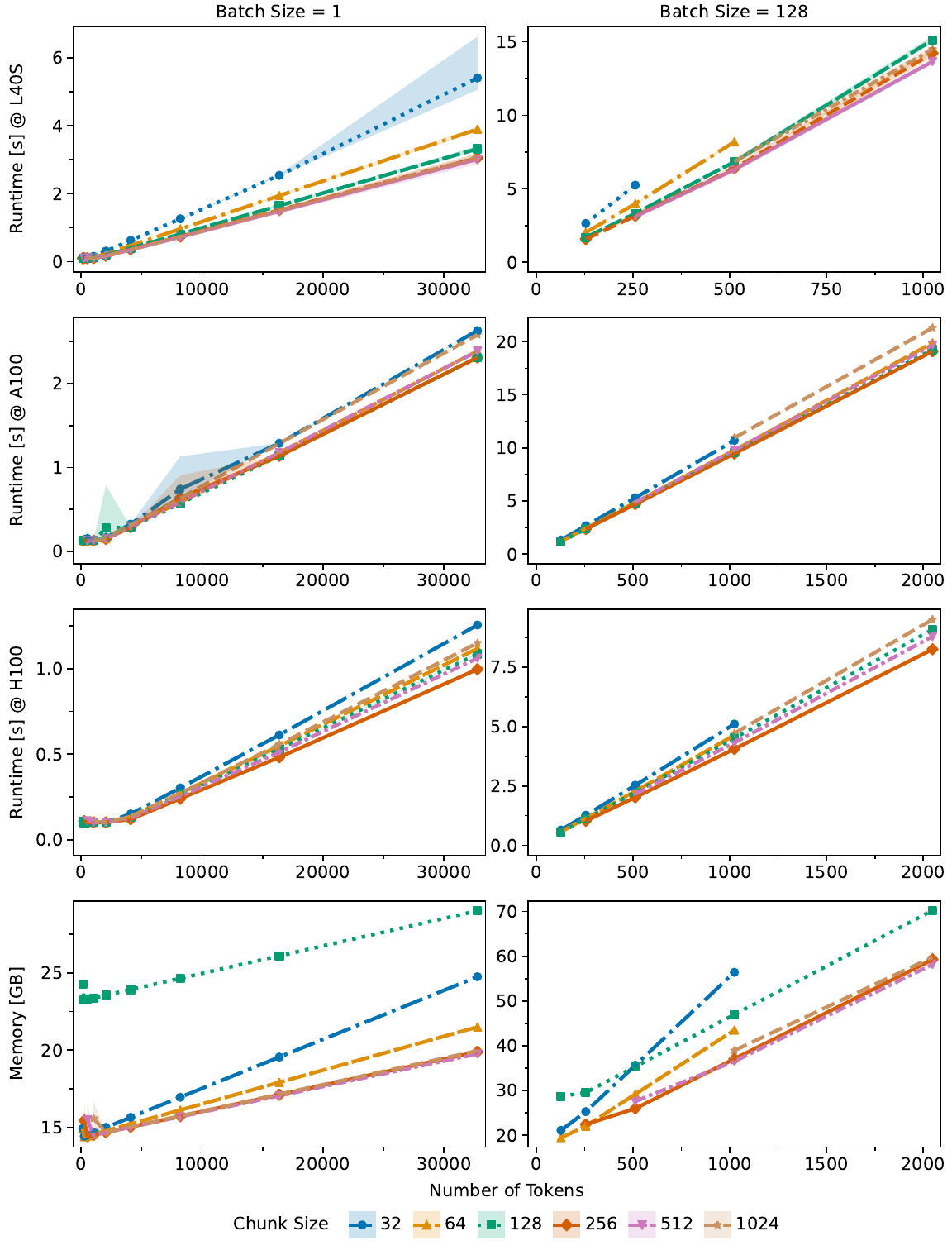}
  \caption{Runtime of \modelcodestral\ using fully horizontal chunked inference with varying chunk sizes for batch sizes 1 and 128, measured on NVIDIA L40S, A100, and H100. Memory usage is consistent across devices and is therefore shown once. Values are reported as mean, minimum, and maximum over three runs.}
  \label{fig:inference-chunk-sizes-all-gpu}
\end{figure*}

\begin{figure*}[t]
  \includegraphics[width=\linewidth]{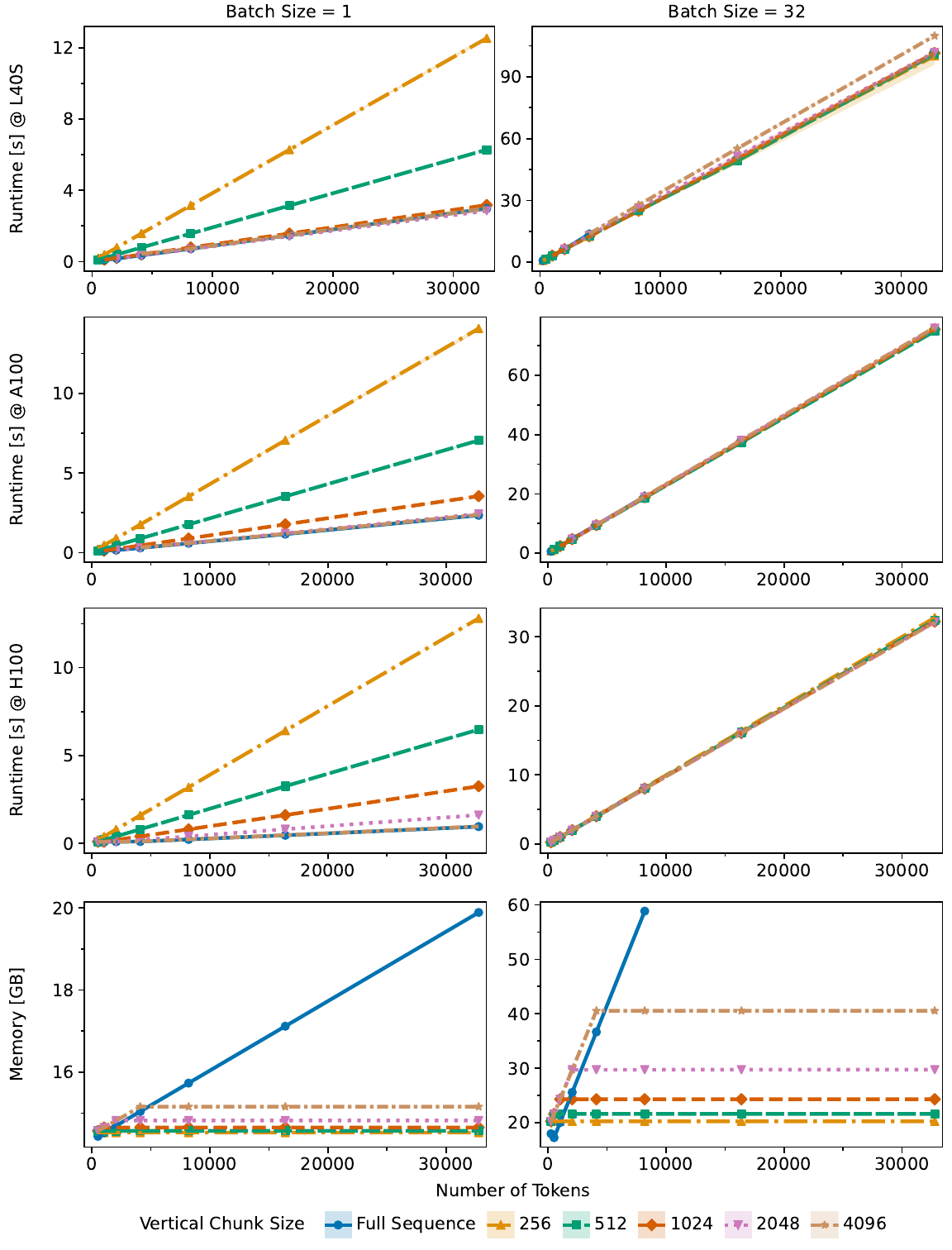}
  \caption{Runtime and memory usage of \modelcodestral\ using vertically chunked inference with a fixed intra-layer chunk size of 256 and varying vertical chunk sizes for batch sizes 1 and 32, measured on NVIDIA L40S, A100, and H100. Memory usage is consistent across devices and is therefore shown once. Values are reported as mean, minimum, and maximum over three runs.}
  \label{fig:inference-vertical-chunk-sizes-all-gpu}
\end{figure*}

\begin{figure*}[t]
  \includegraphics[width=\linewidth]{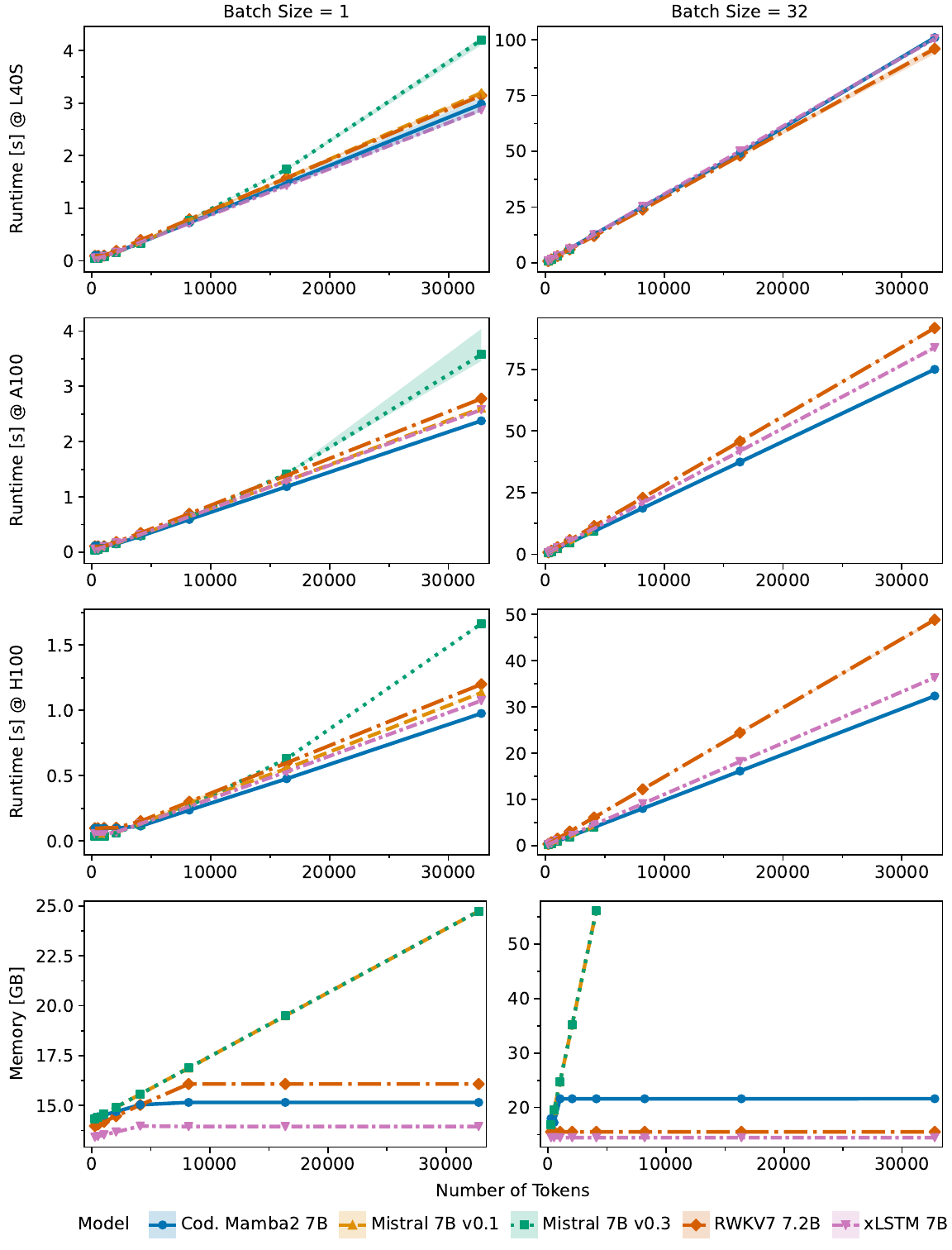}
  \caption{Runtime and memory usage of \modelcodestral, \modelrwkv, and \modelxlstm\ compared to \modelmistralversion{v0.1} and \version{v0.3}, measured on NVIDIA L40S, A100, and H100. Recurrent models use vertically chunked inference: \modelcodestral\ with intra-layer chunk size~256 and vertical chunk sizes~512 (batch~32) and~4096 (batch 1), \modelrwkv\ with chunk size~32 and vertical chunk sizes~64 (batch~32) and~4096 (batch 1), and \modelxlstm\ with chunk size~64 and vertical chunk sizes~128 (batch~32) and 4096 (batch~1). Transformer-based models use FlashAttention-2. Latent states are stored across layers only for input lengths exceeding the vertical chunk size. Memory usage is consistent across devices and is therefore shown once. Values are reported as mean, minimum, and maximum over three runs.}
  \label{fig:inference-model-comparison-all-gpu}
\end{figure*}

To verify that the inference behavior of recurrent models generalizes across hardware, we repeat the key experiments on three accelerators: NVIDIA L40S, A100, and H100. The results, shown in Figures~\ref{fig:inference-chunk-sizes-all-gpu}, \ref{fig:inference-vertical-chunk-sizes-all-gpu}, and \ref{fig:inference-model-comparison-all-gpu}, confirm that the findings from Section~\ref{sec:experiments:inference-evaluation} hold consistently across platforms:
\begin{itemize}
    \item 
    The default chunk size configuration of $Q = 256$ remains optimal for \modelcodestral.
    \item
    The parallelization potential saturates at a vertical chunk size of at most $V = 4096$ for single-sequence inference and at $V$ close to the horizontal chunk size $Q$ for larger batches.
    \item
    The runtime and memory advantages of recurrent models over transformers for longer sequences are preserved regardless of the hardware choice.
\end{itemize}

\subsection{Practical Parameter Selection}
\label{app:parameter-selection}

In our experiments, we found that the default intra-layer chunk size $Q$ works well without modification for all evaluated models. For the vertical chunk size $V$, the appropriate choice depends primarily on the batch size. For the batch sizes larger than one, parallelization nearly saturates at $V = Q$ and fully saturates at $V = 2Q$. This yields runtimes equivalent to full-sequence parallelization, all while maintaining constant activation memory beyond $V$ tokens. Single-sequence inference (batch size 1) requires larger vertical chunk sizes of $V \approx 4096$ to fully saturate the parallelization potential, with the exact saturation point depending on the target hardware.

In practice, we recommend using the model's native chunk size for $Q$ and setting $V = 2Q$ for batched inference or $V = Q$ if memory is constrained. For single-sequence inference, setting $V$ to the largest value that fits in memory yields the best performance for the available memory budget. These settings generalize across the three accelerators (L40S, A100, H100) and three architectures (Mamba2, RWKV, xLSTM) we evaluated.

\section{Training Details}
\label{app:training-details}

\subsection{Training Data}

\begin{table*}[t]
  \centering
  \begin{tabular}{lrrr}
\toprule
Dataset & Total & Empty Text (Count) & Empty Text (Percentage) \\
\midrule
Short Short & 19932 & 1995 & $10.01$ \% \\
STS & 99791 & 3819 & $3.83$ \% \\
Bitext & 89611 & 3047 & $3.40$ \% \\
Short Long & 153934 & 4635 & $3.01$ \% \\
Long Short & 108487 & 2111 & $1.95$ \% \\
Long Long & 19236 & 119 & $0.62$ \% \\
Public Datasets (11) & 1248378 & 0 & $0.00$ \% \\
\midrule
 & 1739369 & 15726 & $0.90$ \% \\
\bottomrule
\end{tabular}

  \caption{Distribution of triples with at least one empty text in the query, positive, or negative example.}
  \label{tab:training-data-empty-strings}
\end{table*}

To compare recurrent and transformer-based embedding models in a controlled setting, we follow the training procedure of \modelefive~\citep{wang-etal-2024-improving-text} as closely as possible. We fine-tuned all models on a combination of the public datasets used to train \modelefive\ and a replication of the synthetically generated training data from~\citet{springer-etal-2025-understanding}. We found that the synthetic data contains almost $16{,}000$ triples where the query, the positive, or the negative example is empty (after removing leading and trailing whitespace), as summarized in Table~\ref{tab:training-data-empty-strings}. To avoid issues during training, we filter out these triples. The instructions used during fine-tuning are provided in Appendix~\ref{app:instructions-training}.

\subsection{Training Configuration}

We trained all models for one epoch using a batch size of 2048, a 100-step learning rate warm-up with linear decay, weight decay of $0.1$, and a loss temperature $\temperature = 0.02$. Following \citet{wang-etal-2024-improving-text} and \citet{springer-etal-2025-understanding}, we limited the maximum sequence length to 512 tokens during training. Lacking explicit sampling strategy details, we construct batches from a single source dataset, with a sampling probability proportional to dataset size. With the exception of \modelmamba, we use the learning rate of $\expnumber{4}{-4}$ reported by \citet{springer-etal-2025-understanding} for all models. We reduced memory consumption during training by using QLoRA \citep{dettmers_qlora_2023} with 4-bit quantization, \texttt{bfloat16} activations, and LoRA \citep{hu_lora_2022} adapters ($r = 16$, $\alpha = 16$) applied to all linear model layers.

For the inference evaluation in Section~\ref{sec:experiments:inference-evaluation}, we additionally fine-tune \modelmistralversion{v0.3} alongside \version{v0.1}. Relative to \version{v0.1}, version \version{v0.2} increased the context window from 8{,}192 to 32{,}768 tokens, adjusted positional encoding parameters, and replaced sliding-window attention with full attention. Version \version{v0.3} inherits these changes and additionally introduces an extended vocabulary.

\subsection{Learning Rate Optimization}

\begin{table*}[t]
  \begin{minipage}{\columnwidth}
    \centering
    \begin{tabular}{rr}
\toprule
Learning Rate & Mean (Task) \\
\midrule
$\expnumber{1.6}{-3}$ & 62.89 \\
$\expnumber{2.2}{-3}$ & 63.19 \\
$\mathbf{\expnumber{2.4}{-3}}$ & \textbf{63.33} \\
$\expnumber{2.6}{-3}$ & 62.77 \\
$\expnumber{2.8}{-3}$ & 62.53 \\
\bottomrule
\end{tabular}

    \caption{Learning rate tuning results for \modelmamba, evaluated on \englishmteb\ after fine-tuning on a subset of the full training data for one epoch.}
    \label{tab:mamba2-learning-rate-tuning}
  \end{minipage}
  \hfill
  \begin{minipage}{\columnwidth}
    \centering
    \begin{tabular}{lrr}
\toprule
Dataset & Samples & Unique Instructions \\
\midrule
Short-Short & 19932 & 12165 \\
Short-Long & 153934 & 96872 \\
Long-Short & 108487 & 67217 \\
Long-Long & 19236 & 11724 \\
\bottomrule
\end{tabular}

    \caption{For datasets used during fine-tuning that contain multiple distinct instructions, the number of samples and unique instructions per dataset are shown.}
    \label{tab:training-instructions-many}
  \end{minipage}
\end{table*}

\modelmamba\ required a considerably higher learning rate to achieve competitive performance. To determine a suitable value, we fine-tuned \modelmamba\ on $\frac{1}{16}$ of the total training data for one epoch. We reduced the batch size by the same factor to a value of 128, ensuring an equal number of gradient updates as in the full training setup. We evaluated a range of learning rates and report the mean score over all tasks of the \englishmteb\ benchmark for the quantized model in Table~\ref{tab:mamba2-learning-rate-tuning}. Performance peaks at a learning rate of $\expnumber{2.4}{-3}$, which we use for the final training run.

\section{Instructions}
\label{app:instructions}

\subsection{Training Instructions}
\label{app:instructions-training}

\begingroup
\renewcommand\tabularxcolumn[1]{m{#1}}
\begin{table*}[t]
  \centering
  \begin{tabularx}{\linewidth}{lX}
\toprule
Datasets & Instruction \\
\midrule
ELI5 & Provided a user question, retrieve the highest voted answers on Reddit ELI5 forum \\
HotpotQA & Given a multi-hop question, retrieve documents that can help answer the question \\
MIRACL & Given a question, retrieve Wikipedia passages that answer the question \\
\multirow{2}{*}{MS MARCO} & Given a web search query, retrieve relevant documents that answer the query \\
 & Given a web search query, retrieve relevant passages that answer the query \\
MrTyDi & Given a question, retrieve Wikipedia passages that answer the question \\
\multirow{2}{*}{NLI} & Given a premise, retrieve a hypothesis that is entailed by the premise \\
 & Retrieve semantically similar text \\
NQ & Given a question, retrieve Wikipedia passages that answer the question \\
\multirow{2}{*}{Quora Duplicates} & Find questions that have the same meaning as the input question \\
 & Given a question, retrieve questions that are semantically equivalent to the given question \\
SQuAD & Retrieve Wikipedia passages that answer the question \\
Synthetic Bitext & Retrieve parallel sentences. \\
Synthetic STS & Retrieve semantically similar text. \\
T2Ranking & Given a Chinese search query, retrieve web passages that answer the question \\
\bottomrule
\end{tabularx}

  \caption{Instructions used during fine-tuning on datasets with one or two distinct instructions.}
  \label{tab:training-instructions-few}
\end{table*}
\endgroup

Table~\ref{tab:training-instructions-many} provides statistics for datasets with multiple instructions, whereas Table~\ref{tab:training-instructions-few} lists the fine-tuning instructions for datasets with one or two distinct instructions.

\subsection{Evaluation Instructions}
\label{app:instructions-evaluation}

\begingroup
\renewcommand\tabularxcolumn[1]{m{#1}}
\begin{table*}[t]
    \centering
    \begin{tabularx}{\linewidth}{lX}
\toprule
Task Type & Instruction \\
\midrule
STS & Retrieve semantically similar text. \\
Summarization & Given a news summary, retrieve other semantically similar summaries \\
BitextMining & Retrieve parallel sentences. \\
\bottomrule
\end{tabularx}

    \caption{Instructions used for the \mteb\ evaluation per task type.}
    \label{tab:mteb-instructions-type}
\end{table*}
\endgroup

\begingroup
\renewcommand\tabularxcolumn[1]{m{#1}}
\setlength{\tabcolsep}{3pt}
\begin{table*}[t]
  \fontsize{9}{12}\selectfont
  \centering
  \begin{tabularx}{\linewidth}{lX}
\toprule
Task & Instruction \\
\midrule
AmazonCounterfactualClassification & Classify a given Amazon customer review text as either counterfactual or not-counterfactual \\
AmazonPolarityClassification & Classify Amazon reviews into positive or negative sentiment \\
AmazonReviewsClassification & Classify the given Amazon review into its appropriate rating category \\
ArguAna & Given a claim, find documents that refute the claim \\
Arxiv(Hierarchical)ClusteringP2P & Identify the main and secondary category of Arxiv papers based on the titles and abstracts \\
Arxiv(Hierarchical)ClusteringS2S & Identify the main and secondary category of Arxiv papers based on the titles \\
AskUbuntuDupQuestions & Retrieve duplicate questions from AskUbuntu forum \\
Banking77Classification & Given an online banking query, find the corresponding intents \\
BiorxivClusteringP2P(.v2) & Identify the main category of Biorxiv papers based on the titles and abstracts \\
BiorxivClusteringS2S & Identify the main category of Biorxiv papers based on the titles \\
\makecell[l]{CQADupstackGamingRetrieval \\ CQADupstackUnixRetrieval} & Given a question, retrieve detailed question descriptions from Stackexchange that are duplicates to the given question \\
ClimateFEVER(HardNegatives) & Given a claim about climate change, retrieve documents that support or refute the claim \\
CovidRetrieval & Given a question on COVID-19, retrieve news articles that answer the question \\
DBPedia & Given a query, retrieve relevant entity descriptions from DBPedia \\
EmotionClassification & Classify the emotion expressed in the given Twitter message into one of the six emotions: anger, fear, joy, love, sadness, and surprise \\
FEVER(HardNegatives) & Given a claim, retrieve documents that support or refute the claim \\
FiQA2018 & Given a financial question, retrieve user replies that best answer the question \\
HotpotQA(HardNegatives) & Given a multi-hop question, retrieve documents that can help answer the question \\
ImdbClassification & Classify the sentiment expressed in the given movie review text from the IMDB dataset \\
\makecell[l]{MIRACLRetrievalHardNegatives \\ NQ} & Given a question, retrieve Wikipedia passages that answer the question \\
MSMARCO & Given a web search query, retrieve relevant passages that answer the query \\
MTOPDomainClassification & Classify the intent domain of the given utterance in task-oriented conversation \\
MTOPIntentClassification & Classify the intent of the given utterance in task-oriented conversation \\
MassiveIntentClassification & Given a user utterance as a query, find the user intents \\
MassiveScenarioClassification & Given a user utterance as a query, find the user scenarios \\
MedrxivClusteringP2P(.v2) & Identify the main category of Medrxiv papers based on the titles and abstracts \\
MedrxivClusteringS2S(.v2) & Identify the main category of Medrxiv papers based on the titles \\
MindSmallReranking & Retrieve relevant news articles based on user browsing history \\
NFCorpus & Given a question, retrieve relevant documents that best answer the question \\
QuoraRetrieval & Given a question, retrieve questions that are semantically equivalent to the given question \\
RedditClustering & Identify the topic or theme of Reddit posts based on the titles \\
RedditClusteringP2P & Identify the topic or theme of Reddit posts based on the titles and posts \\
SCIDOCS & Given a scientific paper title, retrieve paper abstracts that are cited by the given paper \\
SciDocsRR & Given a title of a scientific paper, retrieve the titles of other relevant papers \\
SciFact & Given a scientific claim, retrieve documents that support or refute the claim \\
SprintDuplicateQuestions & Retrieve duplicate questions from Sprint forum \\
StackExchangeClustering(.v2) & Identify the topic or theme of StackExchange posts based on the titles \\
StackExchangeClusteringP2P(.v2) & Identify the topic or theme of StackExchange posts based on the given paragraphs \\
StackOverflowDupQuestions & Retrieve duplicate questions from StackOverflow forum \\
T2Reranking & Given a Chinese search query, retrieve web passages that answer the question \\
TRECCOVID & Given a query on COVID-19, retrieve documents that answer the query \\
Touche2020(Retrieval.v3) & Given a question, retrieve detailed and persuasive arguments that answer the question \\
ToxicConversationsClassification & Classify the given comments as either toxic or not toxic \\
TweetSentimentExtractionClassification & Classify the sentiment of a given tweet as either positive, negative, or neutral \\
TwentyNewsgroupsClustering(.v2) & Identify the topic or theme of the given news articles \\
\makecell[l]{TwitterSemEval2015 \\ TwitterURLCorpus} & Retrieve tweets that are semantically similar to the given tweet \\
\bottomrule
\end{tabularx}

  \vspace{-0.5em}
  \caption{Instructions used for the \mteb\ evaluation per task.}
  \label{tab:mteb-instructions-task}
\end{table*}
\endgroup

We use task-type-specific instructions (Table~\ref{tab:mteb-instructions-type}) and task-specific instructions (Table~\ref{tab:mteb-instructions-task}). For retrieval and reranking tasks, we prepend the instruction to the query only; for all other tasks, we prepend it to all sequences.

\end{document}